\batchmode
\makeatletter
\def\input@path{{\string"C:/Users/Yuan Zhou/Dropbox/YuanHyperspectral/document/GMM/\string"}}
\makeatother
\documentclass[twocolumn,english]{IEEEtran}
\usepackage[T1]{fontenc}
\usepackage[latin9]{inputenc}
\usepackage{babel}
\usepackage{array}
\usepackage{rotating}
\usepackage{mathtools}
\usepackage{url}
\usepackage{multirow}
\usepackage{amsmath}
\usepackage{amsthm}
\usepackage{amssymb}
\usepackage{graphicx}
\usepackage[unicode=true]
 {hyperref}

\makeatletter

\providecommand{\tabularnewline}{\\}

 \let\oldforeign@language\foreign@language
 \DeclareRobustCommand{\foreign@language}[1]{%
   \lowercase{\oldforeign@language{#1}}}
\theoremstyle{plain}
\newtheorem{thm}{\protect\theoremname}

\usepackage[caption=false,font=footnotesize]{subfig}
\usepackage{threeparttable}

\makeatother

\providecommand{\theoremname}{Theorem}

\begin{document}

\title{A Gaussian mixture model representation of endmember variability
in hyperspectral unmixing}

\author{Yuan~Zhou,~\IEEEmembership{Student Member,~IEEE,} Anand Rangarajan,~\IEEEmembership{Member,~IEEE,}
and~Paul~D.~Gader,~\IEEEmembership{Fellow,~IEEE}\thanks{The authors are with the Department of Computer and Information Science
and Engineering, University of Florida, Gainesville, FL, USA. E-mail:
\protect\href{mailto:{yuan,anand,pgader}@cise.ufl.edu}{{yuan,anand,pgader}@cise.ufl.edu}.
This paper has supplementary downloadable material available at http://ieeexplore.ieee.org.,
provided by the author. The material includes a proof of Theorem 2
in the paper. Contact zhouyuanzxcv@gmail.com for further questions
about this work.}}

\markboth{To appear in the IEEE Transactions on Image Processing}{Y. Zhou \MakeLowercase{\emph{et al.}}: A Gaussian mixture model
representation of endmember variability in hyperspectral unmixing}

\IEEEpubid{0000\textendash 0000/00\$00.00~\copyright~2018 IEEE}
\maketitle
\begin{abstract}
Hyperspectral unmixing while considering endmember variability is
usually performed by the normal compositional model (NCM), where the
endmembers for each pixel are assumed to be sampled from unimodal
Gaussian distributions. However, in real applications, the distribution
of a material is often not Gaussian. In this paper, we use Gaussian
mixture models (GMM) to represent endmember variability. We show,
given the GMM starting premise, that the distribution of the mixed
pixel (under the linear mixing model) is also a GMM (and this is shown
from two perspectives). The first perspective originates from the
random variable transformation and gives a conditional density function
of the pixels given the abundances and GMM parameters. With proper
smoothness and sparsity prior constraints on the abundances, the conditional
density function leads to a standard maximum \emph{a posteriori }(MAP\emph{)}
problem which can be solved using generalized expectation maximization.
The second perspective originates from marginalizing over the endmembers
in the GMM, which provides us with a foundation to solve for the endmembers
at each pixel. Hence, compared to the other distribution based methods,
our model can not only estimate the abundances and distribution parameters,
but also the distinct endmember set for each pixel. We tested the
proposed GMM on several synthetic and real datasets, and showed its
potential by comparing it to current popular methods.
\end{abstract}


\section{Introduction}

\IEEEpubidadjcol

\IEEEPARstart{T}{he} formation of hyperspectral images can be simplified
by the \emph{linear mixing model} (LMM), which assumes that the physical
region corresponding to a pixel contains several pure materials, so
that each material contributes a fraction of its spectra based on
area to the final spectra of the pixel. Hence, the observed spectra
$\mathbf{y}_{n}\in\mathbb{R}^{B}$, $n=1,\dots,N$ ($B$ is the number
of wavelengths and $N$ is the number of pixels) is a (non-negative)
linear combination of the pure material (called \emph{endmember})
spectra $\mathbf{m}_{j}\in\mathbb{R}^{B}$, $j=1,\dots,M$ ($M$ is
the number of endmembers), i.e.

\begin{equation}
\mathbf{y}_{n}=\sum_{j=1}^{M}\mathbf{m}_{j}\alpha_{nj}+\mathbf{n}_{n},\,\text{s.t.}\,\alpha_{nj}\ge0,\,\sum_{j=1}^{M}\alpha_{nj}=1,\label{eq:LMM-1}
\end{equation}
where $\alpha_{nj}$ is the proportion (called \emph{abundance}) for
the $j$th endmember at the $n$th pixel (with the positivity and
sum-to-one constraint) and $\mathbf{n}_{n}\in\mathbb{R}^{B}$ is additive
noise. Here, the endmember set $\left\{ \mathbf{m}_{j}:\,j=1,\dots,M\right\} $
is fixed for all the pixels. This model simplifies the unmixing problem
to a matrix factorization one, leading to efficient computation and
simple algorithms such as iterative constrained endmembers (ICE),
vertex component analysis (VCA), piecewise convex multiple-model endmember
detection (PCOMMEND) \cite{berman2004ice,nascimento2005vertex,zare2013piecewise}
etc., which receive comprehensive reviews in \cite{bioucas2012hyperspectral,keshava2002spectral}.

However, in practice the LMM may not be valid in many real scenarios.
Even for a \emph{pure} pixel that only contains one material, its
spectra may not be consistent over the whole image. This is due to
several factors such as atmospheric conditions, topography and intrinsic
variability. For example, in vegetation, multiple scattering and biotic
variation (e.g. differences in biochemistry and water content) cause
different reflectances among the same species. For urban scenes, the
incidence and emergence angles could be different for the same roof,
causing different reflectances. For minerals, the spectroscopy model
developed by Hapke also considers the porosity and roughness of the
material as variable \cite{hapke1981bidirectional}.

In the first and third example above, Eq.~\eqref{eq:LMM-1} can be
generalized to a more abstract form $\mathbf{y}_{n}=F\left(\left\{ \mathbf{m}_{j},\alpha_{nj}:\,j=1,\dots M\right\} \right)$,
which leads to \emph{nonlinear mixing models}. For example, in \cite{halimi2011nonlinear}
the authors used bilinear models to handle the vegetation case, which
was also investigated using several different nonlinear functions
\cite{somers2009nonlinear}. In \cite{heylen2014nonlinear}, the Hapke
model was used to model intimate interaction among minerals. There
are also works that use kernels for flexible nonlinear mixing \cite{broadwater2010generalized,broadwater2007kernel}.
A panoply of nonlinear models can be found in the review article \cite{HeylenParenteGader2014}.
We note that in these models, a fixed endmember set is still assumed
while using a more complicated unmixing model.

\IEEEpubidadjcol

While nonlinear models abound lately, it is still difficult to account
for all the scenarios. On the contrary, the LMM still has physical
significance with the intuitive area assumption. To model real scenarios
more accurately, researchers have taken another route by generalizing
Eq.~\eqref{eq:LMM-1} to 
\begin{equation}
\mathbf{y}_{n}=\sum_{j=1}^{M}\mathbf{m}_{nj}\alpha_{nj}+\mathbf{n}_{n},\label{eq:LMM_n}
\end{equation}
where $\left\{ \mathbf{m}_{nj}\in\mathbb{R}^{B}:\,j=1,\dots,M\right\} ,\,n=1,\dots,N$
could be different for each $n$, i.e. the endmember spectra for each
pixel could be different. This is called \emph{endmember variability},
and has also received a lot of attention in the community \cite{somers2011endmember,zare2014endmember}.
Note that given $\left\{ \mathbf{y}_{n}\right\} $, inferring $\left\{ \mathbf{m}_{nj},\alpha_{nj}\right\} $
is a much more difficult problem than inferring $\left\{ \mathbf{m}_{j},\alpha_{nj}\right\} $
in Eq.~\eqref{eq:LMM-1}. Hence, in many papers $\left\{ \mathbf{m}_{nj}\right\} $
are assumed to be from a spectral library, which is usually called
\emph{supervised unmixing} \cite{du2014spatial,zare2010pce,roberts1998mapping}.
On the other hand, if the endmember spectra are to be extracted from
the image, we call them \emph{unsupervised unmixing} models \cite{halimi2015unsupervised,eches2010bayesian,bateson2000endmember}.
Obviously, unsupervised unmixing is more challenging than its supervised
counterpart and hence more assumptions are used in this case, such
as the spatial smoothness of abundances and endmember variability
\cite{drumetz2016blind,thouvenin2016hyperspectral,halimi2016hyperspectral},
small mutual distance between the endmembers \cite{thouvenin2016hyperspectral},
small magnitude or spectral smoothness of the endmember variability
\cite{thouvenin2016hyperspectral,halimi2016hyperspectral}.

We can also categorize the papers on endmember variability by how
this variability is modeled. In the review paper \cite{zare2014endmember},
it can be modeled as a endmember set \cite{bateson2000endmember,roberts1998mapping}
or as a distribution \cite{zhangpso,eches2010estimating,song2005spectral}.
One of the widely used set based methods is multiple endmember spectral
mixture analysis (MESMA) \cite{roberts1998mapping}, which tries every
endmember combination and selects the one with the smallest error.
There are many variations to the original MESMA. For example, the
multiple-endmember linear spectral unmixing model (MELSUM) solves
the linear equations directly using the pseudo-inverse and discards
the solutions with negative abundances \cite{combe2008analysis};
automatic Monte Carlo unmixing (AutoMCU) picks random combinations
for unmixing and averages the resulting abundances as the final results
\cite{asner2000biogeophysical,asner2002spectral}. Besides MESMA variants,
there are also many other set based methods. For example, endmember
bundles form bundles from automated extracted endmembers, take minimum
and maximum abundances from bundle based unmixing, and average them
as final abundances \cite{bateson2000endmember}; sparse unmixing
imposes a sparsity constraint on the abundances based on endmembers
composed of all spectra from the spectral library \cite{castrodad2011learning}.
A comprehensive review can be found in \cite{somers2011endmember,zare2014endmember}.
One disadvantage of set based methods is that their complexity increases
exponentially with increasing library size hence in practice a laborious
library reduction approach may be required \cite{wetherley2017mapping}. 

The distribution based approaches assume that the endmembers for each
pixel are sampled from probability distributions {[}e.g. Gaussian,
a.k.a. \emph{normal compositional model} (NCM){]}, and hence embrace
large libraries while being numerically tractable \cite{du2014spatial,stein2003application}.
Here, we give an overview of NCM because of its simplicity and popularity
\cite{eches2010bayesian,halimi2015unsupervised,zare2010pce}. Suppose
the \emph{j}th endmember at the \emph{n}th pixel follows a Gaussian
distribution $p\left(\mathbf{m}_{nj}\right)=\mathcal{N}\left(\mathbf{m}_{nj}\vert\boldsymbol{\mu}_{j},\boldsymbol{\Sigma}_{j}\right)$
where $\boldsymbol{\mu}_{j}\in\mathbb{R}^{B}$ and $\boldsymbol{\Sigma}_{j}\in\mathbb{R}^{B\times B}$,
and the additive noise also follows a Gaussian distribution $p\left(\mathbf{n}_{n}\right)=\mathcal{N}\left(\mathbf{n}_{n}|\mathbf{0},\mathbf{D}\right)$
where $\mathbf{D}$ is the noise covariance matrix. The random variable
transformation (r.v.t.) \eqref{eq:LMM_n} suggests that the probability
density function of $\mathbf{y}_{n}$ can be derived as

\begin{equation}
p\left(\mathbf{y}_{n}\vert\boldsymbol{\alpha}_{n},\boldsymbol{\Theta},\mathbf{D}\right)=\mathcal{N}\left(\mathbf{y}_{n}\vert\sum_{j=1}^{M}\alpha_{nj}\boldsymbol{\mu}_{j},\sum_{j=1}^{M}\alpha_{nj}^{2}\boldsymbol{\Sigma}_{j}+\mathbf{D}\right),\label{eq:ncm_yn}
\end{equation}
where \begin{small}$\boldsymbol{\alpha}_{n}\vcentcolon=\left[\alpha_{n1},\dots,\alpha_{nM}\right]^{T}$,
$\boldsymbol{\Theta}\vcentcolon=\left\{ \boldsymbol{\mu}_{j},\boldsymbol{\Sigma}_{j}:\,j=1,\dots,M\right\} $\end{small}.
The conditional density function in \eqref{eq:ncm_yn} is usually
embedded in a Bayesian framework such that we can incorporate priors
and also estimate hyperparameters. Then, NCM uses different optimization
approaches, e.g. expectation maximization \cite{stein2003application},
sampling methods \cite{eches2010bayesian,eches2010estimating,halimi2015unsupervised},
particle swarm optimization \cite{zhangpso}, to determine the parameters
$\left\{ \boldsymbol{\mu}_{j},\boldsymbol{\Sigma}_{j}\right\} $ and
$\left\{ \alpha_{nj}\right\} $.

\begin{figure}
\begin{centering}
\includegraphics[width=8.5cm]{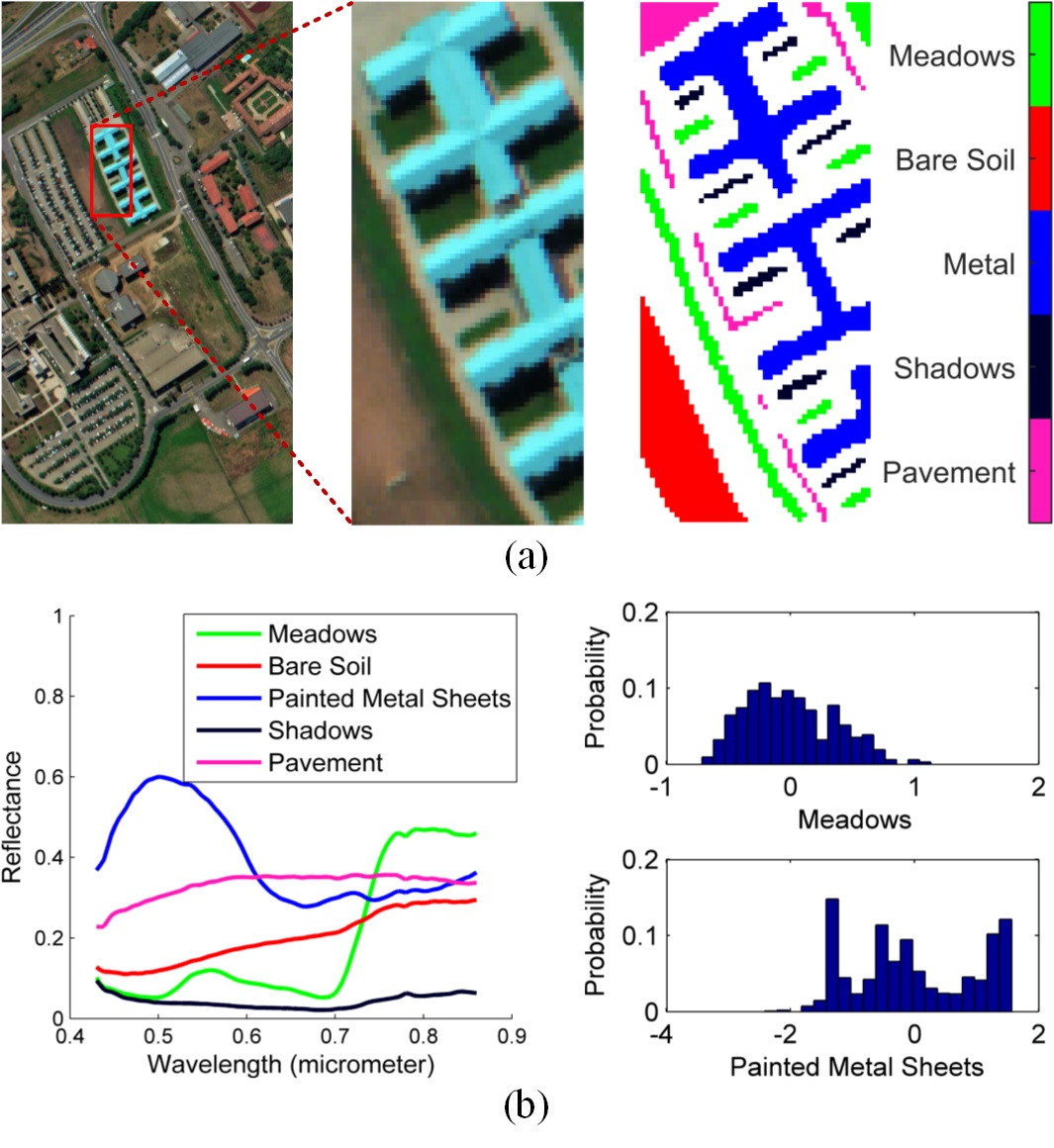}
\par\end{centering}
\caption{(a) Original Pavia University image and selected ROI with its ground
truth image. (b) Mean spectra of the identified 5 endmembers and histograms
of meadows and painted metal sheets (shadow is termed as endmember
to conform with the LMM though the area under shadow can be any material).
PCA is used to project the multidimensional pixels to single values
which are counted in the histograms. Although the histogram of meadows
may appear to be a Gaussian distribution, that of painted metal sheets
is obviously neither a unimodal Gaussian or Beta distribution.}

\label{fig:pavia_roi=000026hist}
\end{figure}

There are few papers that use other distributions. In \cite{du2014spatial},
Xiaoxiao Du \emph{et al.} note that the Gaussian distribution may
allow negative values which are not realistic. In addition, the real
distribution may be skewed. Hence, they introduce a Beta compositional
model (BCM) to model the variability. The problem is that the true
distribution may not be well approximated by any unimodal distribution.
Consider the Pavia University dataset shown in Fig.~\ref{fig:pavia_roi=000026hist},
where the multidimensional pixels are projected to one dimension to
afford better visualization. Among the manually identified materials,
we can see that although the histogram of meadows may look like a
Gaussian distribution, that of painted metal sheets has multiple peaks
and cannot be approximated by either a Gaussian or Beta distribution.
This is due to different angles of these sheets on the roof. Since
each piece of metal sheet is tilted, it forms a cluster of reflectances
which contributes to a peak in the histogram. This example shows that
we should use a more flexible distribution to represent the endmember
variability.

\begin{figure*}
\begin{centering}
\includegraphics[width=18cm]{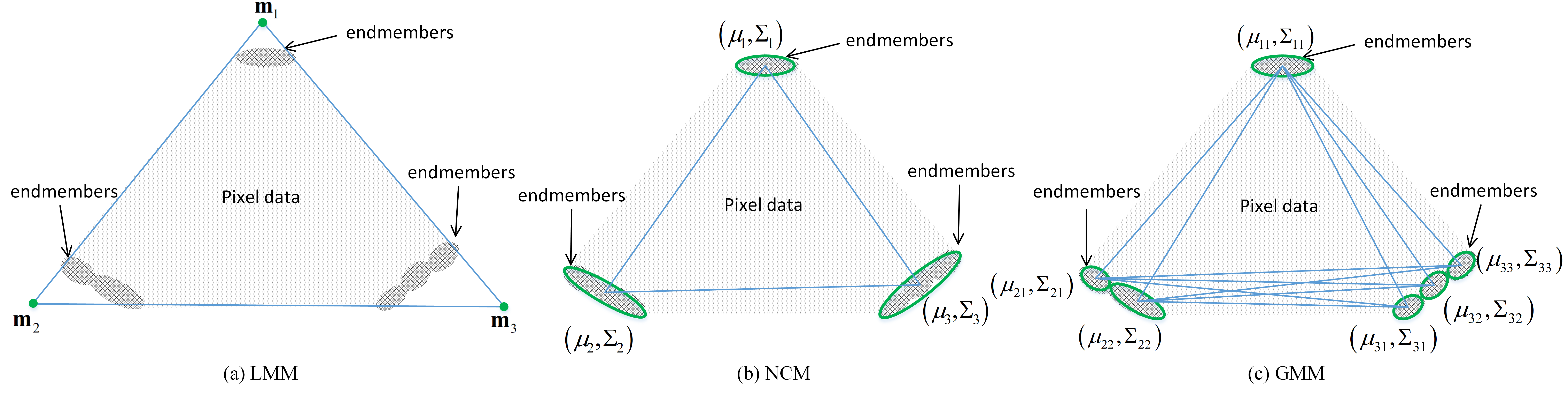}
\par\end{centering}
\caption{Comparison of the mechanisms among LMM, NCM and GMM. We have 3 endmembers
represented by the darken gray areas. LMM tries to find a set of endmembers
that fit the pixel data. NCM tries to find a set of Gaussian centers
that fit the pixel data, with error weighted by the covariance matrices.
GMM tries to find Gaussian centers such that all their linear combinations
fit the pixel data, with each weighted by the prior $\pi_{\mathbf{k}}$.
We may use 6 endmembers with NCM, but then the prior information is
lost.}

\label{fig:LMM_NCM_GMM}
\end{figure*}

In this paper, we use a mixture of Gaussians to approximate any distribution
that an endmember may exhibit, and solve the LMM by considering endmember
variability. In a nutshell, the Gaussian mixture model (GMM) models
$p\left(\mathbf{m}_{nj}\right)$ by a mixture of Gaussians, say $p\left(\mathbf{m}_{nj}\right)=\sum_{k}\pi_{jk}\mathcal{N}\left(\mathbf{m}_{nj}|\boldsymbol{\mu}_{jk},\boldsymbol{\Sigma}_{jk}\right)$,
and then obtains the distribution of $\mathbf{y}_{n}$ by the r.v.t.
\eqref{eq:LMM_n}, which turns out to be another mixture of Gaussians
and can be used for inference of the unknown parameters. Here, we
briefly explain how GMM works intuitively by comparing it to the NCM
with the details given later. The maximum likelihood estimate (MLE)
of NCM (using \eqref{eq:ncm_yn}) aims to find $\left\{ \boldsymbol{\mu}_{j}\right\} $
such that its linear combination matches $\mathbf{y}_{n}$. Contrary
to NCM, GMM aims to find $\left\{ \boldsymbol{\mu}_{jk}\right\} $
such that \emph{all} of its linear combinations match $\mathbf{y}_{n}$.
Suppose we have $\boldsymbol{\mu}_{11}$, $\boldsymbol{\mu}_{21}$,
$\boldsymbol{\mu}_{22}$, $\boldsymbol{\mu}_{31}$, $\boldsymbol{\mu}_{32}$,
$\boldsymbol{\mu}_{33}$: then there are 6 combinations as explained
in Fig.~\ref{fig:LMM_NCM_GMM}, but with emphasis weighted by $\left\{ \pi_{jk}\right\} $
which determines the prior probability of each linear combination.

Based on the GMM formulation, we propose a supervised version and
an unsupervised version for unmixing. The supervised version takes
a library as input and estimates the abundances. The unsupervised
version assumes that there are regions of pure pixels, hence segments
the image first to get pure pixels and then performs unmixing. Another
advantage over the other distribution based methods is that we can
also estimate the endmembers for each pixel, which is not achievable
by NCM or BCM. Note that estimating endmembers for each pixel is generally
common in non-distribution methods, both from the signal processing
community \cite{thouvenin2016hyperspectral,drumetz2016blind,halimi2016hyperspectral}
or the remote sensing community \cite{roberts1998mapping,combe2008analysis}.
But it is often achieved in the context of least-squares based unmixing
\cite{tits2012potential,iordache2014dynamic,tits2014site}, unlike
what we propose here using distribution based unmixing.

\textbf{Notation}: As usual, $\mathcal{N}\left(\mathbf{x}\vert\boldsymbol{\mu},\boldsymbol{\Sigma}\right)$
denotes the multivariate Gaussian density function with center $\boldsymbol{\mu}$
and covariance matrix $\boldsymbol{\Sigma}$. Let $\mathbf{A}\in\mathbb{R}^{m\times n}$
be a matrix with $m$ rows and $n$ columns. The Hadamard product
of two matrices (elementwise multiplication) is denoted by $\circ$
while the Kronecker product is denoted by $\otimes$. $\left(\mathbf{A}\right)_{jk}$
denotes the element at the \emph{j}th row and \emph{k}th column of
matrix $\mathbf{A}$. $\left(\mathbf{A}\right)_{j}$ denotes the \emph{j}th
row of $\mathbf{A}$ transposed (treating $\mathbf{A}$ as a vector),
i.e. for $\mathbf{A}=\left[\mathbf{a}_{1},\dots\mathbf{a}_{n}\right]^{T}$,
$\left(\mathbf{A}\right)_{j}=\mathbf{a}_{j}$. $\text{vec}\left(\mathbf{A}\right)$
denotes the vectorization of $\mathbf{A}$, i.e. concatenating the
columns of $\mathbf{A}$. $\delta_{jk}=1$ when $j=k$ and 0 otherwise.
$\mathbb{E}_{\mathbf{x}}\left(f\left(\mathbf{x}\right)\right)$ is
the expected value of $f\left(\mathbf{x}\right)$ given random variable
$\mathbf{x}$. We use $i=\sqrt{-1}$ instead of as an index throughout
the paper.

\section{Mathematical Preliminaries\label{sec:Mathematical-Preliminary}}

\subsection{Linear combination of GMM random variables}

To use the Gaussian mixture model to model endmember variability,
we start by assuming that $\mathbf{m}_{nj}$ follows a Gaussian mixture
model (GMM) and the noise also follows a Gaussian distribution. The
distribution of $\mathbf{y}_{n}$ is obtained using the following
theorem.
\begin{thm}
\label{thm:linear_comb_GMM}If the random variable $\mathbf{m}_{nj}$
has a density function
\begin{equation}
p\left(\mathbf{m}_{nj}\vert\boldsymbol{\Theta}\right)\vcentcolon=f_{\mathbf{m}_{j}}\left(\mathbf{m}_{nj}\right)=\sum_{k=1}^{K_{j}}\pi_{jk}\mathcal{N}\left(\mathbf{m}_{nj}\vert\boldsymbol{\mu}_{jk},\boldsymbol{\Sigma}_{jk}\right),\label{eq:gmm}
\end{equation}
s.t. $\pi_{jk}\ge0,\,\sum_{k=1}^{K_{j}}\pi_{jk}=1$, with $K_{j}$
being the number of components, $\pi_{jk}$ ($\boldsymbol{\mu}_{jk}\in\mathbb{R}^{B}$
or $\boldsymbol{\Sigma}_{jk}\in\mathbb{R}^{B\times B}$) being the
weight (mean or covariance matrix) of its kth Gaussian component,
$\boldsymbol{\Theta}\vcentcolon=\left\{ \pi_{jk},\boldsymbol{\mu}_{jk},\boldsymbol{\Sigma}_{jk}:\,j=1,\dots,M,k=1,\dots,K_{j}\right\} $,
$\left\{ \mathbf{m}_{nj}:\,j=1,\dots,M\right\} $ are independent,
and the random variable $\mathbf{n}_{n}$ has a density function $p\left(\mathbf{n}_{n}\right):=\mathcal{N}\left(\mathbf{n}_{n}\vert\mathbf{0},\mathbf{D}\right)$,
then the density function of $\mathbf{y}_{n}$ given by the r.v.t.
$\mathbf{y}_{n}=\sum_{j=1}^{M}\mathbf{m}_{nj}\alpha_{nj}+\mathbf{n}_{n}$
is another GMM
\begin{equation}
p\left(\mathbf{y}_{n}\vert\boldsymbol{\alpha}_{n},\boldsymbol{\Theta},\mathbf{D}\right)=\sum_{\mathbf{k}\in\mathcal{K}}\pi_{\mathbf{k}}\mathcal{N}\left(\mathbf{y}_{n}\vert\boldsymbol{\mu}_{n\mathbf{k}},\boldsymbol{\Sigma}_{n\mathbf{k}}\right),\label{eq:density_y_theta}
\end{equation}
where $\mathcal{K}\vcentcolon=\left\{ 1,\dots,K_{1}\right\} \times\left\{ 1,\dots,K_{2}\right\} \times\cdots\times\left\{ 1,\dots,K_{M}\right\} $
is the Cartesian product of the $M$ index sets, $\mathbf{k}\vcentcolon=\left(k_{1},\dots k_{M}\right)\in\mathcal{K}$,
$\pi_{\mathbf{k}}\in\mathbb{R}$, $\boldsymbol{\mu}_{n\mathbf{k}}\in\mathbb{R}^{B}$,
$\boldsymbol{\Sigma}_{n\mathbf{k}}\in\mathbb{R}^{B\times B}$ are
defined by
\begin{equation}
\pi_{\mathbf{k}}:=\prod_{j=1}^{M}\pi_{jk_{j}},\,\boldsymbol{\mu}_{n\mathbf{k}}\vcentcolon=\sum_{j=1}^{M}\alpha_{nj}\boldsymbol{\mu}_{jk_{j}},\,\boldsymbol{\Sigma}_{n\mathbf{k}}\vcentcolon=\sum_{j=1}^{M}\alpha_{nj}^{2}\boldsymbol{\Sigma}_{jk_{j}}+\mathbf{D}.\label{eq:mu_Sigma_nk}
\end{equation}
\end{thm}
The proof is detailed using a characteristic function (c.f.) approach.

We first consider the distribution of the intermediate variable $\mathbf{z}_{n}=\sum_{j=1}^{M}\mathbf{m}_{nj}\alpha_{nj}$.
The c.f. of $f_{\mathbf{m}_{j}}$ in \eqref{eq:gmm}, $\phi_{\mathbf{m}_{j}}\left(\mathbf{t}\right):\,\mathbb{R}^{B}\rightarrow\mathbb{C}$,
is given by
\begin{alignat}{1}
\phi_{\mathbf{m}_{j}}\left(\mathbf{t}\right) & =\mathbb{E}_{\mathbf{m}_{j}}\left(e^{i\mathbf{t}^{T}\mathbf{x}}\right)=\int_{\mathbb{R}^{B}}e^{i\mathbf{t}^{T}\mathbf{x}}f_{\mathbf{m}_{j}}\left(\mathbf{x}\right)d\mathbf{x}\nonumber \\
 & =\sum_{k=1}^{K_{j}}\pi_{jk}\int_{\mathbb{R}^{B}}e^{i\mathbf{t}^{T}\mathbf{x}}\mathcal{N}\left(\mathbf{x}\vert\boldsymbol{\mu}_{jk},\boldsymbol{\Sigma}_{jk}\right)d\mathbf{x}\nonumber \\
 & =\sum_{k=1}^{K_{j}}\pi_{jk}\phi_{jk}\left(\mathbf{t}\right),\label{eq:cf_mj}
\end{alignat}
where $\phi_{jk}\left(\mathbf{t}\right)$ denotes the c.f. of the
Gaussian distribution $\mathcal{N}\left(\mathbf{x}\vert\boldsymbol{\mu}_{jk},\boldsymbol{\Sigma}_{jk}\right)$
as

\begin{equation}
\phi_{jk}\left(\mathbf{t}\right)\vcentcolon=\exp\left(i\mathbf{t}^{T}\boldsymbol{\mu}_{jk}-\frac{1}{2}\mathbf{t}^{T}\boldsymbol{\Sigma}_{jk}\mathbf{t}\right).\label{eq:cf_normal}
\end{equation}
Assuming $\mathbf{m}_{n1},\dots,\mathbf{m}_{nM}$ are independent,
we can obtain the c.f. of the linear combination of these $\mathbf{m}_{nj}$
by multiplying \eqref{eq:cf_mj} as 
\begin{alignat*}{1}
 & \phi_{\mathbf{z}_{n}}\left(\mathbf{t}\right)=\phi_{\mathbf{m}_{n1}\alpha_{n1}+\cdots+\mathbf{m}_{nM}\alpha_{nM}}\left(\mathbf{t}\right)=\prod_{j=1}^{M}\phi_{\mathbf{m}_{j}}\left(\alpha_{nj}\mathbf{t}\right)\\
= & \sum_{k_{1}=1}^{K_{1}}\cdots\sum_{k_{M}=1}^{K_{M}}\pi_{1k_{1}}\cdots\pi_{Mk_{M}}\phi_{1k_{1}}\left(\alpha_{n1}\mathbf{t}\right)\cdots\phi_{Mk_{M}}\left(\alpha_{nM}\mathbf{t}\right).
\end{alignat*}
Let $\mathcal{K}$, $\mathbf{k}$, $\pi_{\mathbf{k}}$ be defined
as in Theorem~\ref{thm:linear_comb_GMM}. We can write the above
multiple summations in an elegant way:
\begin{equation}
\phi_{\mathbf{z}_{n}}\left(\mathbf{t}\right)=\sum_{\mathbf{k}\in\mathcal{K}}\pi_{\mathbf{k}}\phi_{n\mathbf{k}}\left(\mathbf{t}\right),\label{eq:cf_z}
\end{equation}
where $\pi_{\mathbf{k}}\ge0$, $\sum_{\mathbf{k}\in\mathcal{K}}\pi_{\mathbf{k}}=1$
and 
\begin{alignat*}{1}
 & \phi_{n\mathbf{k}}\left(\mathbf{t}\right)\vcentcolon=\phi_{1k_{1}}\left(\alpha_{n1}\mathbf{t}\right)\cdots\phi_{Mk_{M}}\left(\alpha_{nM}\mathbf{t}\right)\\
= & \exp\left\{ i\mathbf{t}^{T}\left(\sum_{j=1}^{M}\alpha_{nj}\boldsymbol{\mu}_{jk_{j}}\right)-\frac{1}{2}\mathbf{t}^{T}\left(\sum_{j=1}^{M}\alpha_{nj}^{2}\boldsymbol{\Sigma}_{jk_{j}}\right)\mathbf{t}\right\} ,
\end{alignat*}
where \eqref{eq:cf_normal} is used. Since $\phi_{n\mathbf{k}}\left(\mathbf{t}\right)$
also has a form of c.f. of a Gaussian distribution, the corresponding
distribution turns out to be $\mathcal{N}\left(\mathbf{x}\vert\sum_{j}\alpha_{nj}\boldsymbol{\mu}_{jk_{j}},\sum_{j}\alpha_{nj}^{2}\boldsymbol{\Sigma}_{jk_{j}}\right)$.
Hence, the distribution of $\mathbf{z}_{n}$ can be obtained by the
Fourier transform of \eqref{eq:cf_z}
\begin{alignat}{1}
f_{\mathbf{z}_{n}}\left(\mathbf{z}_{n}\right) & =\frac{1}{\left(2\pi\right)^{B}}\int_{\mathbb{R}^{B}}e^{-i\mathbf{t}^{T}\mathbf{z}_{n}}\phi_{\mathbf{z}_{n}}\left(\mathbf{t}\right)d\mathbf{t}\nonumber \\
 & =\frac{1}{\left(2\pi\right)^{B}}\int_{\mathbb{R}^{B}}e^{-i\mathbf{t}^{T}\mathbf{z}_{n}}\sum_{\mathbf{k}\in\mathcal{K}}\pi_{\mathbf{k}}\phi_{n\mathbf{k}}\left(\mathbf{t}\right)d\mathbf{t}\nonumber \\
 & =\sum_{\mathbf{k}\in\mathcal{K}}\pi_{\mathbf{k}}\mathcal{N}\left(\mathbf{z}_{n}\vert\sum_{j=1}^{M}\alpha_{nj}\boldsymbol{\mu}_{jk_{j}},\sum_{j=1}^{M}\alpha_{nj}^{2}\boldsymbol{\Sigma}_{jk_{j}}\right),\label{eq:density_z}
\end{alignat}
which is still a mixture of Gaussians.

After finding the distribution of the linear combination, we can add
the noise term to find the distribution of $\mathbf{y}_{n}$. Suppose
the noise also follows a Gaussian distribution, $p\left(\mathbf{n}_{n}\right)\vcentcolon=f_{\mathbf{n}_{n}}\left(\mathbf{n}_{n}\right)=\mathcal{N}\left(\mathbf{n}_{n}\vert\mathbf{0},\mathbf{D}\right),$
where $\mathbf{D}$ is the noise covariance matrix. We assume that
the noise at different wavelengths is independent ($\sigma_{k}^{2}$
being the noise variance of the $k$th band), i.e. $\mathbf{D}=\text{diag}\left(\sigma_{1}^{2},\sigma_{2}^{2},\dots,\sigma_{B}^{2}\right)\in\mathbb{R}^{B\times B}$
(if it is not independent, the noise can actually be easily whitened
to be independent as in \cite{lee1990enhancement}). Its c.f. has
the following form 
\begin{equation}
\phi_{\mathbf{n}_{n}}\left(\mathbf{t}\right)=\exp\left(-\frac{1}{2}\mathbf{t}^{T}\mathbf{D}\mathbf{t}\right)\label{eq:cf_n}
\end{equation}
by \eqref{eq:cf_normal}. Then the c.f. of $\mathbf{y}_{n}$ can be
obtained by multiplying \eqref{eq:cf_z} and \eqref{eq:cf_n} (as
$\mathbf{z}_{n}$ and $\mathbf{n}_{n}$ are independent)
\begin{alignat*}{1}
\phi_{\mathbf{y}_{n}}\left(\mathbf{t}\right) & =\phi_{\mathbf{z}_{n}}\left(\mathbf{t}\right)\phi_{\mathbf{n}_{n}}\left(\mathbf{t}\right)=\sum_{\mathbf{k}\in\mathcal{K}}\pi_{\mathbf{k}}\phi_{\mathbf{n}_{n}}\left(\mathbf{t}\right)\phi_{n\mathbf{k}}\left(\mathbf{t}\right)\\
 & =\sum_{\mathbf{k}\in\mathcal{K}}\pi_{\mathbf{k}}\exp\left\{ i\mathbf{t}^{T}\boldsymbol{\mu}_{n\mathbf{k}}-\frac{1}{2}\mathbf{t}^{T}\boldsymbol{\Sigma}_{n\mathbf{k}}\mathbf{t}\right\} ,
\end{alignat*}
where $\boldsymbol{\mu}_{n\mathbf{k}}$ and $\boldsymbol{\Sigma}_{n\mathbf{k}}$
are defined in \eqref{eq:mu_Sigma_nk}. Finally, the distribution
of $\mathbf{y}$ can be shown to be \eqref{eq:density_y_theta} by
the Fourier transform again as in \eqref{eq:density_z}.

If $\mathcal{K}=\left\{ 1\right\} \times\left\{ 1\right\} \times\cdots\times\left\{ 1\right\} $,
i.e. each endmember has only one Gaussian component, we have $\pi_{11}=1,\dots,\pi_{M1}=1$,
then $\pi_{\mathbf{k}}=\pi_{11}\cdots\pi_{M1}=1$. The distribution
of $\mathbf{y}_{n}$ becomes
\begin{equation}
p\left(\mathbf{y}_{n}\vert\boldsymbol{\alpha}_{n},\boldsymbol{\Theta},\mathbf{D}\right)=\mathcal{N}\left(\mathbf{y}_{n}\vert\sum_{j=1}^{M}\alpha_{nj}\boldsymbol{\mu}_{j1},\sum_{j=1}^{M}\alpha_{nj}^{2}\boldsymbol{\Sigma}_{j1}+\mathbf{D}\right),\label{eq:density_y_single}
\end{equation}
which is exactly the NCM in \eqref{eq:ncm_yn}.

\subsection{Another perspective}

Theorem~\ref{thm:linear_comb_GMM} obtains the density of each pixel
by directly performing a r.v.t. based on the LMM, which can be used
to estimate the abundances and distribution parameters. Here, we will
obtain the density from another perspective, which provides a foundation
to estimate the endmembers for each pixel. Again, let the noise follow
the density function $p\left(\mathbf{n}_{n}\right):=\mathcal{N}\left(\mathbf{n}_{n}\vert\mathbf{0},\mathbf{D}\right)$.
Considering $\left\{ \mathbf{m}_{nj}\right\} $ and $\left\{ \alpha_{nj}\right\} $
as fixed values, the r.v.t. $\mathbf{y}_{n}=\sum_{j}\mathbf{m}_{nj}\alpha_{nj}+\mathbf{n}_{n}$
implies that the density of $\mathbf{y}_{n}$ is given by
\begin{equation}
p\left(\mathbf{y}_{n}|\boldsymbol{\alpha}_{n},\mathbf{M}_{n},\mathbf{D}\right)=\mathcal{N}\left(\mathbf{y}_{n}\vert\sum_{j}\mathbf{m}_{nj}\alpha_{nj},\mathbf{D}\right)\label{eq:alternate_density_yn}
\end{equation}
where $\mathbf{M}_{n}=\left[\mathbf{m}_{n1},\dots,\mathbf{m}_{nM}\right]^{T}\in\mathbb{R}^{M\times B}$
are the endmembers for the \emph{n}th pixel. We have the following
theorem which gives the same result as in Theorem~\ref{thm:linear_comb_GMM}. 
\begin{thm}
\label{thm:linear_comb_GMM2}If the random variables $\left\{ \mathbf{m}_{nj}:j=1,\dots,M\right\} $
follow GMM distributions
\[
p\left(\mathbf{m}_{nj}\vert\boldsymbol{\Theta}\right):=\sum_{k=1}^{K_{j}}\pi_{jk}\mathcal{N}\left(\mathbf{m}_{nj}\vert\boldsymbol{\mu}_{jk},\boldsymbol{\Sigma}_{jk}\right),
\]
and they are independent, i.e. 
\begin{equation}
p\left(\mathbf{M}_{n}\vert\boldsymbol{\Theta}\right)=\prod_{j=1}^{M}p\left(\mathbf{m}_{nj}\vert\boldsymbol{\Theta}\right),\label{eq:density_Mn}
\end{equation}
then the conditional density $p\left(\mathbf{y}_{n}\vert\boldsymbol{\alpha}_{n},\boldsymbol{\Theta},\mathbf{D}\right)$
obtained by marginalizing $\mathbf{M}_{n}$ in $p\left(\mathbf{y}_{n},\mathbf{M}_{n}\vert\boldsymbol{\alpha}_{n},\boldsymbol{\Theta},\mathbf{D}\right)$
has the same form as in Theorem~\ref{thm:linear_comb_GMM}:
\begin{align*}
p\left(\mathbf{y}_{n}\vert\boldsymbol{\alpha}_{n},\boldsymbol{\Theta},\mathbf{D}\right) & =\int p\left(\mathbf{y}_{n}\vert\boldsymbol{\alpha}_{n},\mathbf{M}_{n},\mathbf{D}\right)p\left(\mathbf{M}_{n}\vert\boldsymbol{\Theta}\right)d\mathbf{M}_{n}\\
 & =\sum_{\mathbf{k}\in\mathcal{K}}\pi_{\mathbf{k}}\mathcal{N}\left(\mathbf{y}_{n}\vert\boldsymbol{\mu}_{n\mathbf{k}},\boldsymbol{\Sigma}_{n\mathbf{k}}\right),
\end{align*}
where $p\left(\mathbf{y}_{n}|\boldsymbol{\alpha}_{n},\mathbf{M}_{n},\mathbf{D}\right)=\mathcal{N}\left(\mathbf{y}_{n}\vert\sum_{j}\mathbf{m}_{nj}\alpha_{nj},\mathbf{D}\right)$.
\end{thm}
The proof is much more complicated (in terms of algebra) and therefore
relegated to the supplemental material of the paper. 

\subsection{An example\label{subsec:An-example}}

We give an example to illustrate the basic idea of this paper. Suppose
we have $M=4$ endmembers with $K_{1}=1$, $K_{2}=2$, $K_{3}=3$,
$K_{4}=1$. Their distributions follow \eqref{eq:gmm} with $\boldsymbol{\mu}_{jk},\boldsymbol{\Sigma}_{jk}$,
$j=1,2,3,4$, $k=1,...,K_{j}$. Let the weights of these components
be $\pi_{11}=\pi_{41}=1$, $\pi_{21}=0.3$, $\pi_{22}=0.7$, $\pi_{31}=0.2$,
$\pi_{32}=0.4$, $\pi_{33}=0.4$. Then, $\mathcal{K}$ has 6 entries
from the Cartesian product, $\left\{ 1\right\} \times\left\{ 1,2\right\} \times\left\{ 1,2,3\right\} \times\left\{ 1\right\} $.
We list the values for $\pi_{\mathbf{k}}$, $\boldsymbol{\mu}_{n\mathbf{k}}$
in Table~\ref{table:example}. For example, for $\mathbf{k}=\left(1,2,3,1\right)$,
$\pi_{\mathbf{k}}=\pi_{11}\pi_{22}\pi_{33}\pi_{41}=0.28$. The value
of $\boldsymbol{\mu}_{n\mathbf{k}}$ is a linear combination of $\boldsymbol{\mu}_{jk}$
(pick one component for each $j$) based on the configuration $\mathbf{k}$.
Hence, the distribution of $\mathbf{y}_{n}$ in \eqref{eq:density_y_theta}
is a Gaussian mixture of 6 components with $\pi_{\mathbf{k}}$, $\boldsymbol{\mu}_{n\mathbf{k}}$
given in Table~\ref{table:example} ($\boldsymbol{\Sigma}_{n\mathbf{k}}$
can be derived similar to $\boldsymbol{\mu}_{n\mathbf{k}}$). Recalling
the intuition in Fig.~\ref{fig:LMM_NCM_GMM}, we will show that applying
it to hyperspectral unmixing will force each pixel to match all the
$\boldsymbol{\mu}_{n\mathbf{k}}$s, but with emphasis determined by
$\pi_{n\mathbf{k}}$.

\begin{table}
\caption{Values for the various quantities in the simple example.}

\begin{centering}
\begin{tabular}{|c|c|c|}
\hline 
$\mathbf{k}$ &
$\pi_{\mathbf{k}}$ &
$\boldsymbol{\mu}_{n\mathbf{k}}$ in \eqref{eq:mu_Sigma_nk}\tabularnewline
\hline 
\hline 
$\left(1,1,1,1\right)$ &
0.06 &
$\alpha_{n1}\boldsymbol{\mu}_{11}+\alpha_{n2}\boldsymbol{\mu}_{21}+\alpha_{n3}\boldsymbol{\mu}_{31}+\alpha_{n4}\boldsymbol{\mu}_{41}$\tabularnewline
\hline 
$\left(1,2,1,1\right)$ &
0.14 &
$\alpha_{n1}\boldsymbol{\mu}_{11}+\alpha_{n2}\boldsymbol{\mu}_{22}+\alpha_{n3}\boldsymbol{\mu}_{31}+\alpha_{n4}\boldsymbol{\mu}_{41}$\tabularnewline
\hline 
$\left(1,1,2,1\right)$ &
0.12 &
$\alpha_{n1}\boldsymbol{\mu}_{11}+\alpha_{n2}\boldsymbol{\mu}_{21}+\alpha_{n3}\boldsymbol{\mu}_{32}+\alpha_{n4}\boldsymbol{\mu}_{41}$\tabularnewline
\hline 
$\left(1,2,2,1\right)$ &
0.28 &
$\alpha_{n1}\boldsymbol{\mu}_{11}+\alpha_{n2}\boldsymbol{\mu}_{22}+\alpha_{n3}\boldsymbol{\mu}_{32}+\alpha_{n4}\boldsymbol{\mu}_{41}$\tabularnewline
\hline 
$\left(1,1,3,1\right)$ &
0.12 &
$\alpha_{n1}\boldsymbol{\mu}_{11}+\alpha_{n2}\boldsymbol{\mu}_{21}+\alpha_{n3}\boldsymbol{\mu}_{33}+\alpha_{n4}\boldsymbol{\mu}_{41}$\tabularnewline
\hline 
$\left(1,2,3,1\right)$ &
0.28 &
$\alpha_{n1}\boldsymbol{\mu}_{11}+\alpha_{n2}\boldsymbol{\mu}_{22}+\alpha_{n3}\boldsymbol{\mu}_{33}+\alpha_{n4}\boldsymbol{\mu}_{41}$\tabularnewline
\hline 
\end{tabular}
\par\end{centering}
\label{table:example}
\end{table}

\section{Gaussian Mixture Model for Endmember Variability}

\subsection{The GMM for hyperspectral unmixing\label{subsec:The-GMM-for}}

Based on the analysis in Section~\ref{sec:Mathematical-Preliminary},
we can model the conditional distribution of all the pixels $\mathbf{Y}\vcentcolon=\left[\mathbf{y}_{1},\dots,\mathbf{y}_{N}\right]^{T}\in\mathbb{R}^{N\times B}$
given all the abundances $\mathbf{A}\vcentcolon=\left[\boldsymbol{\alpha}_{1},\dots,\boldsymbol{\alpha}_{N}\right]^{T}\in\mathbb{R}^{N\times M}$
($\boldsymbol{\alpha}_{n}\vcentcolon=\left[\alpha_{n1},\dots,\alpha_{nM}\right]^{T}$)
and GMM parameters, which leads to a maximum\emph{ a posteriori} (MAP)
problem. Using the result in \eqref{eq:density_y_theta} and assuming
the conditional distributions of $\mathbf{y}_{n}$ are independent,
the distribution of $\mathbf{Y}$ given $\mathbf{A},\boldsymbol{\Theta},\mathbf{D}$
becomes
\begin{equation}
p\left(\mathbf{Y}\vert\mathbf{A},\boldsymbol{\Theta},\mathbf{D}\right)=\prod_{n=1}^{N}p\left(\mathbf{y}_{n}\vert\boldsymbol{\alpha}_{n},\boldsymbol{\Theta},\mathbf{D}\right).\label{eq:density_Y_theta}
\end{equation}
Based on the hyperspectral unmixing context, we can set the priors
for $\mathbf{A}$. Suppose we use the same prior on $\mathbf{A}$
as in \cite{zhou2016spatial}, i.e. 
\begin{eqnarray}
p\left(\mathbf{A}\right) & \propto & \exp\left\{ -\frac{\beta_{1}}{2}\text{Tr}\left(\mathbf{A}^{T}\mathbf{L}\mathbf{A}\right)+\frac{\beta_{2}}{2}\text{Tr}\left(\mathbf{A}^{T}\mathbf{A}\right)\right\} \nonumber \\
 & = & \exp\left\{ -\frac{\beta_{1}}{2}\text{Tr}\left(\mathbf{A}^{T}\mathbf{K}\mathbf{A}\right)\right\} ,\label{eq:APdf}
\end{eqnarray}
where $\mathbf{L}$ is a \emph{graph Laplacian} matrix constructed
from $w_{nm},\,n,m=1,\dots,N$ with $w_{nm}=e^{-\Vert\mathbf{y}_{n}-\mathbf{y}_{m}\Vert^{2}/2B\eta^{2}}$
for neighboring pixels and 0 otherwise. We have $\text{Tr}\left(\mathbf{A}^{T}\mathbf{L}\mathbf{A}\right)=\frac{1}{2}\sum_{n,m}w_{nm}\Vert\boldsymbol{\alpha}_{n}-\boldsymbol{\alpha}_{m}\Vert^{2}$),
$\mathbf{K}=\mathbf{L}-\frac{\beta_{2}}{\beta_{1}}\mathbf{I}_{N}$
(suppose $\beta_{1}\neq0$) with $\beta_{1}$ controlling smoothness
and $\beta_{2}$ controlling sparsity of the abundance maps. 

From the conditional density function and the priors, Bayes' theorem
says the posterior is given by
\begin{equation}
p\left(\mathbf{A},\boldsymbol{\Theta}\vert\mathbf{Y},\mathbf{D}\right)\propto p\left(\mathbf{Y}\vert\mathbf{A},\boldsymbol{\Theta},\mathbf{D}\right)p\left(\mathbf{A}\right)p\left(\boldsymbol{\Theta}\right),\label{eq:posterior}
\end{equation}
where $p\left(\boldsymbol{\Theta}\right)$ is assumed to follow a
uniform distribution. Maximizing $p\left(\mathbf{A},\boldsymbol{\Theta}\vert\mathbf{Y},\mathbf{D}\right)$
is equivalent to minimizing $-\log p\left(\mathbf{A},\boldsymbol{\Theta}\vert\mathbf{Y},\mathbf{D}\right)$,
which reduces to the following form by combining \eqref{eq:density_y_theta},
\eqref{eq:density_Y_theta}, \eqref{eq:APdf} and \eqref{eq:posterior}:
\begin{equation}
\mathcal{E}\left(\mathbf{A},\boldsymbol{\Theta}\right)=-\sum_{n=1}^{N}\log\sum_{\mathbf{k}\in\mathcal{K}}\pi_{\mathbf{k}}\mathcal{N}\left(\mathbf{y}_{n}\vert\boldsymbol{\mu}_{n\mathbf{k}},\boldsymbol{\Sigma}_{n\mathbf{k}}\right)+\mathcal{E}_{\text{prior}}(\mathbf{A}),\label{eq:obj_fun}
\end{equation}
\[
\text{s.t.}\,\pi_{\mathbf{k}}\ge0,\,\sum_{\mathbf{k}\in\mathcal{K}}\pi_{\mathbf{k}}=1,\,\alpha_{nj}\ge0,\,\sum_{j=1}^{M}\alpha_{nj}=1,\,\forall n
\]
where $\mathcal{E}_{\text{prior}}(\mathbf{A})=\frac{\beta_{1}}{2}\text{Tr}\left(\mathbf{A}^{T}\mathbf{K}\mathbf{A}\right)$,
and $\boldsymbol{\mu}_{n\mathbf{k}},\boldsymbol{\Sigma}_{n\mathbf{k}}$
are defined in \eqref{eq:mu_Sigma_nk}.

\subsection{Relationships to least-squares, NCM and MESMA}

Let us focus on the first term in \eqref{eq:obj_fun} and call it
the \emph{data fidelity term}. We can relate it to NCM and the least-squares
term $\sum_{n}\Vert\mathbf{y}_{n}-\sum_{j}\alpha_{nj}\mathbf{m}_{j}\Vert^{2}$
as used in previous research. The data fidelity term in NCM follows
\eqref{eq:ncm_yn} and is based on minimizing the negative log-likelihood
\begin{equation}
-\log p\left(\mathbf{Y}\right)=-\log\prod_{n=1}^{N}p\left(\mathbf{y}_{n}\right)=-\sum_{n=1}^{N}\log\mathcal{N}\left(\mathbf{y}_{n}\vert\boldsymbol{\mu}_{n\mathbf{1}},\boldsymbol{\Sigma}_{n\mathbf{1}}\right)\label{eq:obj_fun_ncm}
\end{equation}
by assuming $\mathbf{y}_{n}$s are independent, where $\boldsymbol{\mu}_{n\mathbf{1}}:=\sum_{j}\alpha_{nj}\boldsymbol{\mu}_{j}$,
$\boldsymbol{\Sigma}_{n\mathbf{1}}:=\sum_{j}\alpha_{nj}^{2}\boldsymbol{\Sigma}_{j}+\sigma^{2}\mathbf{I}_{B}$.
Expanding \eqref{eq:obj_fun_ncm} using the form of the Gaussian distribution
leads to the objective function
\begin{equation}
\sum_{n=1}^{N}\text{log}\left|\boldsymbol{\Sigma}_{n\mathbf{1}}\right|+\sum_{n=1}^{N}\left(\mathbf{y}_{n}-\boldsymbol{\mu}_{n\mathbf{1}}\right)^{T}\boldsymbol{\Sigma}_{n\mathbf{1}}^{-1}\left(\mathbf{y}_{n}-\boldsymbol{\mu}_{n\mathbf{1}}\right).\label{eq:obj_fun_ncm1}
\end{equation}
We can see that the least-squares minimization is a special case of
NCM with $\Vert\boldsymbol{\Sigma}_{j}\Vert_{F}\rightarrow0$, i.e.
when there is little endmember variability.

The proposed GMM further generalizes NCM from a statistical perspective.
Since $\pi_{jk}$ represents the prior probability of the latent variable
in a GMM, $\pi_{\mathbf{k}}$ represents the prior probability of
picking a combination. If we see $\mathbf{k}$ as a (discrete) random
variable whose sample space is $\mathcal{K}$, \eqref{eq:density_y_theta}
can be seen as 
\[
p\left(\mathbf{y}_{n}\vert\boldsymbol{\alpha}_{n},\boldsymbol{\Theta},\mathbf{D}\right)=\sum_{\mathbf{k}\in\mathcal{K}}p\left(\mathbf{k}\right)p\left(\mathbf{y}_{n}\vert\mathbf{k},\boldsymbol{\alpha}_{n},\boldsymbol{\Theta},\mathbf{D}\right),
\]
where $p\left(\mathbf{k}\right)=\pi_{\mathbf{k}}$ and $p\left(\mathbf{y}_{n}\vert\mathbf{k},\boldsymbol{\alpha}_{n},\boldsymbol{\Theta},\mathbf{D}\right)=\mathcal{N}\left(\mathbf{y}_{n}\vert\boldsymbol{\mu}_{n\mathbf{k}},\boldsymbol{\Sigma}_{n\mathbf{k}}\right)$.
From this perspective, each pixel is generated by first sampling $\mathbf{k}$,
then sampling a Gaussian distribution determined by $\mathbf{k},\boldsymbol{\Theta}$.
Unlike NCM that tries to make each $\mathbf{y}_{n}$ close to $\boldsymbol{\mu}_{n\mathbf{1}}$
which is a linear combination of a fixed set $\left\{ \boldsymbol{\mu}_{j}\right\} $,
GMM further generalizes it by trying to make $\mathbf{y}_{n}$ close
to every $\boldsymbol{\mu}_{n\mathbf{k}}$ which are all the possible
linear combinations of $\left\{ \boldsymbol{\mu}_{jk}\right\} $.
It makes sense that the summation in \eqref{eq:obj_fun} is weighted
by $\pi_{\mathbf{k}}$ in a way that if one combination has a high
probability to appear, i.e. $\pi_{\mathbf{k}}$ is larger for a certain
$\mathbf{k}$, the effort is biased to make $\mathbf{y}_{n}$ closer
to this particular $\boldsymbol{\mu}_{n\mathbf{k}}$. Fig.~\ref{fig:LMM_NCM_GMM}
shows the differences among these.

The widely adopted MESMA takes a library of endmember spectra as input,
tries all the combinations and pick the combination with least reconstruction
error. The philosophy is similar to our model despite the fundamental
difference that MESMA is explicit whereas we are implicit in terms
of linear combinations. Compared to MESMA, the GMM approach separates
the library into $M$ groups where each group represents a material
and is clustered into several centers, such that the combination can
only take place by picking one center from each group. Also, the size
of each cluster affects the probability of picking its center. Hence,
our model can adapt to very large library sizes as long as the number
of clusters does not increase too much.

\subsection{Optimization}

Estimating the parameters of GMMs has been studied extensively, from
early expectation maximization (EM) from the statistical community
to projection based clustering from the computer science community
\cite{achlioptas2005spectral,vlassis2002greedy}. There are simple
and deterministic algorithms, which usually require the centers of
Gaussian be separable. However, we face a more challenging problem
since each pixel is generated by a different GMM determined by the
coefficients $\boldsymbol{\alpha}_{n}$. Since EM can be seen as a
special case of Majoriziation-Minimization algorithms \cite{langeOptimization},
which is more flexible, we adopt this approach. Considering that we
have too many parameters $\mathbf{A},\boldsymbol{\Theta}$ to update
in the M step, they are updated sequentially as long as the complete
data log-likelihood increases. This is also called \emph{generalized
expectation maximization} (GEM) \cite{meng1993maximum}.

Following the routine of EM, the E step calculates the posterior probability
of the latent variable given the observed data and old parameters
\begin{equation}
\gamma_{n\mathbf{k}}=\frac{\pi_{\mathbf{k}}\mathcal{N}\left(\mathbf{y}_{n}\vert\boldsymbol{\mu}_{n\mathbf{k}},\boldsymbol{\Sigma}_{n\mathbf{k}}\right)}{\sum_{\mathbf{k}\in\mathcal{K}}\pi_{\mathbf{k}}\mathcal{N}\left(\mathbf{y}_{n}\vert\boldsymbol{\mu}_{n\mathbf{k}},\boldsymbol{\Sigma}_{n\mathbf{k}}\right)}.\label{eq:E-step}
\end{equation}
The M step usually maximizes the expected value of the complete data
log-likelihood. Here, we have priors in the Bayesian formulation.
Hence, we need to minimize
\begin{equation}
\mathcal{E}_{M}=-\sum_{n=1}^{N}\sum_{\mathbf{k}\in\mathcal{K}}\gamma_{n\mathbf{k}}\left\{ \log\pi_{\mathbf{k}}+\log\mathcal{N}\left(\mathbf{y}_{n}\vert\boldsymbol{\mu}_{n\mathbf{k}},\boldsymbol{\Sigma}_{n\mathbf{k}}\right)\right\} +\mathcal{E}_{\text{prior}}.\label{eq:E_M}
\end{equation}
This leads to a common update step for $\pi_{\mathbf{k}}$ as
\begin{equation}
\pi_{\mathbf{k}}=\frac{1}{N}\sum_{n=1}^{N}\gamma_{n\mathbf{k}}.\label{eq:M-step_wk}
\end{equation}
We now focus on updating $\left\{ \boldsymbol{\mu}_{jk},\boldsymbol{\Sigma}_{jk}\right\} $
and $\mathbf{A}$. To achieve this, we require the derivatives of
$\mathcal{E}_{M}$ in \eqref{eq:E_M} w.r.t. $\boldsymbol{\mu}_{jk},\boldsymbol{\Sigma}_{jk},\alpha_{nj}$.
After some tedious algebra using \eqref{eq:mu_Sigma_nk}, we get
\begin{equation}
\frac{\partial\mathcal{E}_{M}}{\partial\boldsymbol{\mu}_{jl}}=-\sum_{n=1}^{N}\sum_{\mathbf{k}\in\mathcal{K}}\delta_{lk_{j}}\alpha_{nj}\boldsymbol{\lambda}_{n\mathbf{k}}\label{eq:M-step_mu_jk}
\end{equation}
\begin{equation}
\frac{\partial\mathcal{E}_{M}}{\partial\boldsymbol{\Sigma}_{jl}}=-\sum_{n=1}^{N}\sum_{\mathbf{k}\in\mathcal{K}}\delta_{lk_{j}}\alpha_{nj}^{2}\boldsymbol{\Psi}_{n\mathbf{k}},\label{eq:M-step_sigma_jk}
\end{equation}
\begin{align}
\frac{\partial\mathcal{E}_{M}}{\partial\alpha_{nj}}= & -\sum_{\mathbf{k}\in\mathcal{K}}\boldsymbol{\lambda}_{n\mathbf{k}}^{T}\boldsymbol{\mu}_{jk_{j}}-2\alpha_{nj}\sum_{\mathbf{k}\in\mathcal{K}}\text{Tr}\left(\boldsymbol{\Psi}_{n\mathbf{k}}^{T}\boldsymbol{\Sigma}_{jk_{j}}\right)\nonumber \\
 & +\beta_{1}\left(\mathbf{K}\mathbf{A}\right)_{nj},\label{eq:M-step_A}
\end{align}
where $\boldsymbol{\lambda}_{n\mathbf{k}}\in\mathbb{R}^{B\times1}$
and $\boldsymbol{\Psi}_{n\mathbf{k}}\in\mathbb{R}^{B\times B}$ are
given by 
\begin{equation}
\boldsymbol{\lambda}_{n\mathbf{k}}=\gamma_{n\mathbf{k}}\boldsymbol{\Sigma}_{n\mathbf{k}}^{-1}\left(\mathbf{y}_{n}-\boldsymbol{\mu}_{n\mathbf{k}}\right),\label{eq:lambda_nk}
\end{equation}
\begin{equation}
\boldsymbol{\Psi}_{n\mathbf{k}}=\frac{1}{2}\gamma_{n\mathbf{k}}\boldsymbol{\Sigma}_{n\mathbf{k}}^{-T}\left(\mathbf{y}_{n}-\boldsymbol{\mu}_{n\mathbf{k}}\right)\left(\mathbf{y}_{n}-\boldsymbol{\mu}_{n\mathbf{k}}\right)^{T}\boldsymbol{\Sigma}_{n\mathbf{k}}^{-T}-\frac{1}{2}\gamma_{n\mathbf{k}}\boldsymbol{\Sigma}_{n\mathbf{k}}^{-T}.\label{eq:psi_nk}
\end{equation}
It is better to represent the derivatives in matrix forms for the
sake of implementation convenience. Considering the multiple summations
in \eqref{eq:M-step_mu_jk}, \eqref{eq:M-step_sigma_jk} and \eqref{eq:M-step_A},
we can write them as

\begin{equation}
\frac{\partial\mathcal{E}_{M}}{\partial\boldsymbol{\mu}_{jl}}=-\sum_{\mathbf{k}\in\mathcal{K}}\delta_{lk_{j}}\left(\mathbf{A}^{T}\boldsymbol{\Lambda}_{\mathbf{k}}\right)_{j},\label{eq:M-step_mu_jk1}
\end{equation}
\begin{equation}
\frac{\partial\mathcal{E}_{M}}{\partial\text{vec}\left(\boldsymbol{\Sigma}_{jl}\right)}=-\sum_{\mathbf{k}\in\mathcal{K}}\delta_{lk_{j}}\left(\left(\mathbf{A}\circ\mathbf{A}\right)^{T}\boldsymbol{\Psi}_{\mathbf{k}}\right)_{j},\label{eq:M-step_sigma_jk1}
\end{equation}
\begin{equation}
\frac{\partial\mathcal{E}_{M}}{\partial\mathbf{A}}=-\sum_{\mathbf{k}\in\mathcal{K}}\boldsymbol{\Lambda}_{\mathbf{k}}\mathbf{R}_{\mathbf{k}}^{T}-2\mathbf{A}\circ\sum_{\mathbf{k}\in\mathcal{K}}\boldsymbol{\Psi}_{\mathbf{k}}\mathbf{S}_{\mathbf{k}}^{T}+\beta_{1}\mathbf{K}\mathbf{A},\label{eq:M-step_A1}
\end{equation}
where $\boldsymbol{\Lambda}_{\mathbf{k}}\in\mathbb{R}^{N\times B}$,
$\boldsymbol{\Psi}_{\mathbf{k}}\in\mathbb{R}^{N\times B^{2}}$ denote
the matrices formed by $\left\{ \boldsymbol{\lambda}_{n\mathbf{k}},\boldsymbol{\Psi}_{n\mathbf{k}}\right\} $
as follows
\[
\boldsymbol{\Lambda}_{\mathbf{k}}:=\left[\boldsymbol{\lambda}_{1\mathbf{k}},\boldsymbol{\lambda}_{2\mathbf{k}},\dots,\boldsymbol{\lambda}_{N\mathbf{k}}\right]^{T},
\]
\[
\boldsymbol{\Psi}_{\mathbf{k}}:=\left[\text{vec}\left(\boldsymbol{\Psi}_{1\mathbf{k}}\right),\text{vec}\left(\boldsymbol{\Psi}_{2\mathbf{k}}\right),\dots,\text{vec}\left(\boldsymbol{\Psi}_{N\mathbf{k}}\right)\right]^{T},
\]
and $\mathbf{R}_{\mathbf{k}}\in\mathbb{R}^{M\times B}$, $\mathbf{S}_{\mathbf{k}}\in\mathbb{R}^{M\times B^{2}}$
are defined by 
\begin{equation}
\mathbf{R}_{\mathbf{k}}:=\left[\boldsymbol{\mu}_{1k_{1}},\boldsymbol{\mu}_{2k_{2}},\dots,\boldsymbol{\mu}_{Mk_{M}}\right]^{T},\label{eq:R_k}
\end{equation}
\begin{equation}
\mathbf{S}_{\mathbf{k}}:=\left[\text{vec}\left(\boldsymbol{\Sigma}_{1k_{1}}\right),\text{vec}\left(\boldsymbol{\Sigma}_{2k_{2}}\right),\dots,\text{vec}\left(\boldsymbol{\Sigma}_{Mk_{M}}\right)\right]^{T}.\label{eq:S_k}
\end{equation}
The minimum of $\mathcal{E}_{M}$ corresponds to $\frac{\partial\mathcal{E}_{M}}{\partial\boldsymbol{\mu}_{jl}}=0$,
$\frac{\partial\mathcal{E}_{M}}{\partial\boldsymbol{\Sigma}_{jl}}=0$,
and $\frac{\partial\mathcal{E}_{M}}{\partial\mathbf{A}}=0$ if the
optimization problem is unconstrained. However, since we have the
non-negativity and sum-to-one constraint to $\alpha_{nj}$ and positive
definite constraint of $\boldsymbol{\Sigma}_{jk}$, minimizing $\mathcal{E}_{M}$
is very difficult. Therefore, in each M step, we only decrease this
objective function by \emph{projected gradient descent} (please see
Section 2.3 in \cite{bertsekas1999nonlinear}, \cite{lin2007projected})
using \eqref{eq:M-step_mu_jk1}, \eqref{eq:M-step_sigma_jk1} and
\eqref{eq:M-step_A1}, where the projection functions for $\mathbf{A}$
and $\left\{ \boldsymbol{\Sigma}_{jk}\right\} $ are the same as in
\cite{zhou2016spatial}.

Finally, from the estimated $\pi_{\mathbf{k}}$, we can recover the
sets of weights as $\pi_{jl}=\sum_{\mathbf{k}\in\mathcal{K}}\delta_{lk_{j}}\pi_{\mathbf{k}}$.

\subsection{Model selection\label{subsec:Model-selection}}

The number of components $K_{j}$ can be specified or estimated from
the data. For the latter case, we have some pure pixels and estimate
$K_{j}$ by deploying a standard model selection method. Suppose we
have $N_{j}$ pure pixels $\mathbf{Y}_{j}:=\left[\mathbf{y}_{1}^{j},\mathbf{y}_{2}^{j},\dots,\mathbf{y}_{N_{j}}^{j}\right]^{T}\in\mathbb{R}^{N_{j}\times B}$
for the \emph{j}th endmember, $f_{\mathbf{m}_{j}}\left(\mathbf{y}\vert\boldsymbol{\Theta}_{j}\right)$
is the estimated density function with $\boldsymbol{\Theta}_{j}\vcentcolon=\left\{ \pi_{jk},\boldsymbol{\mu}_{jk},\boldsymbol{\Sigma}_{jk}:\,k=1,\dots,K_{j}\right\} $,
$g_{\mathbf{m}_{j}}\left(\mathbf{y}\right)$ is the true density function.
The information criterion based model selection approach tries to
find $K_{j}$ that minimizes their difference, e.g. the Kullback-Leibler
(KL) divergence
\begin{eqnarray*}
\mathcal{D}_{\text{KL}}\left(g_{\mathbf{m}_{j}}\Vert f_{\mathbf{m}_{j}}\right) & = & \int_{\mathbb{R}^{B}}g_{\mathbf{m}_{j}}\left(\mathbf{y}\right)\log\frac{g_{\mathbf{m}_{j}}\left(\mathbf{y}\right)}{f_{\mathbf{m}_{j}}\left(\mathbf{y}\vert\boldsymbol{\Theta}_{j}\right)}d\mathbf{y}\\
 & \approx & -\frac{1}{N_{j}}\sum_{n=1}^{N_{j}}\text{log}f_{\mathbf{m}_{j}}\left(\mathbf{y}_{n}^{j}\vert\boldsymbol{\Theta}_{j}\right)+\text{const},
\end{eqnarray*}
where the approximation of $\int g_{\mathbf{m}_{j}}\left(\mathbf{y}\right)\log f_{\mathbf{m}_{j}}\left(\mathbf{y}\vert\boldsymbol{\Theta}_{j}\right)d\mathbf{y}$
by the log-likelihood is usually biased as the empirical distribution
function is closer to the fitted distribution than the true one. Akaike's
information criterion is one way to approximate the bias. Here, we
use the cross-validation-based information criterion (CVIC) to correct
for the bias \cite{mclachlan2014number,smyth2000model}. Let
\begin{equation}
\mathcal{L}_{\mathbf{Y}_{j}}\left(\boldsymbol{\Theta}_{j}\right)=\sum_{n=1}^{N_{j}}\text{log}f_{\mathbf{m}_{j}}\left(\mathbf{y}_{n}^{j}\vert\boldsymbol{\Theta}_{j}\right).\label{eq:model_sel_MLE}
\end{equation}
The \emph{V}-fold cross validation (we use $V=5$ here) divides the
input set $\mathbf{Y}_{j}$ into $V$ subsets $\left\{ \mathbf{Y}_{j}^{1},\mathbf{Y}_{j}^{2},\dots,\mathbf{Y}_{j}^{V}\right\} $
with equal sizes. Then for each subset $\mathbf{Y}_{j}^{v}$, $v=1,\dots,V$,
the remaining data are used to replace $\mathbf{Y}_{j}$ in \eqref{eq:model_sel_MLE}
such that \eqref{eq:model_sel_MLE} is maximized by $\boldsymbol{\Theta}_{j}^{v}$.
Then $\mathcal{L}_{K_{j}}=\sum_{v}\mathcal{L}_{\mathbf{Y}_{j}^{v}}\left(\boldsymbol{\Theta}_{j}^{v}\right)$
is evaluated and the optimal $\hat{K_{j}}=\arg\max_{K_{j}}\mathcal{L}_{K_{j}}$. 

\subsection{Implementation details}

The algorithm can be implemented in a supervised or unsupervised manner.
In both cases, because of the large computational cost, we project
the pixel data to a low dimensional space by principal component analysis
(PCA) and perform the optimization, the result then projected back
to the original space. Let $\mathbf{E}\in\mathbb{R}^{B\times d}$
be the projection matrix and $\mathbf{c}\in\mathbb{R}^{B}$ be the
translation vector, then
\[
\mathbf{E}^{T}\left(\mathbf{y}_{n}-\mathbf{c}\right)=\sum_{j=1}^{M}\mathbf{E}^{T}\left(\mathbf{m}_{nj}-\mathbf{c}\right)\alpha_{nj}+\mathbf{E}^{T}\mathbf{n}_{n}.
\]
This means that for the projected pixels, the $j$th endmember $\mathbf{m}_{nj}^{\prime}=\mathbf{E}^{T}\left(\mathbf{m}_{nj}-\mathbf{c}\right)$
follows a distribution 
\[
p\left(\mathbf{m}_{nj}^{\prime}\vert\boldsymbol{\Theta}\right)=\sum_{k=1}^{K_{j}}\pi_{jk}\mathcal{N}\left(\mathbf{m}_{nj}^{\prime}\vert\mathbf{E}^{T}\left(\boldsymbol{\mu}_{jk}-\mathbf{c}\right),\mathbf{E}^{T}\boldsymbol{\Sigma}_{jk}\mathbf{E}\right)
\]
 and the noise $\mathbf{n}_{n}^{\prime}=\mathbf{E}^{T}\mathbf{n}_{n}$
follows $\mathcal{N}\left(\mathbf{n}_{n}^{\prime}\vert\mathbf{0},\mathbf{E}^{T}\mathbf{D}\mathbf{E}\right)$.

In the supervised unmixing scenario, we assume that a library of endmember
spectra is known. After estimating the number of components following
Section~\ref{subsec:Model-selection}, and calculating $\boldsymbol{\Theta}$
using the standard EM algorithm, we only need to update $\gamma_{n\mathbf{k}}$
by \eqref{eq:E-step} and $\mathbf{A}$ by \eqref{eq:M-step_A1} with
$\pi_{\mathbf{k}}$, $\boldsymbol{\mu}_{jk}$ and $\boldsymbol{\Sigma}_{jk}$
fixed. The initialization of $\mathbf{A}$ can utilize the multiple
combinations of means. For each $\boldsymbol{\alpha}_{n}$, we first
set $\boldsymbol{\alpha}_{n\mathbf{k}}\leftarrow\left(\mathbf{R}_{\mathbf{k}}\mathbf{R}_{\mathbf{k}}^{T}+\epsilon\mathbf{I}_{M}\right)^{-1}\mathbf{R}_{\mathbf{k}}\mathbf{y}_{n}$,
then project it to the simplex space, and finally set $\boldsymbol{\alpha}_{n}\leftarrow\boldsymbol{\alpha}_{n\mathbf{\hat{k}}}$
with $\hat{\mathbf{k}}=\arg\min_{\mathbf{k}}\Vert\mathbf{y}_{n}-\mathbf{R}_{\mathbf{k}}^{T}\boldsymbol{\alpha}_{n\mathbf{k}}\Vert^{2}$,
i.e. choose the $\boldsymbol{\alpha}_{n\mathbf{k}}$ that minimizes
the reconstruction error.

In the unsupervised unmixing scenario, we will assume the resolution
is high enough such that the hyperspectral image can be segmented
into several regions where the interior pixels in each region are
pure pixels. The optimization is performed in several steps, where
we first obtain a segmentation result, then use CVIC to determine
the number of components, and finally estimate $\mathbf{A}$ with
$\boldsymbol{\Theta}$ fixed. The details are given as follows.

Step 1: Initialization. We start with $K_{j}=1,\,\forall j$ and use
K-means to find the initial means $\mathbf{R}_{\mathbf{1}}$. The
initial $\mathbf{A}$ is set to $\mathbf{A}\leftarrow\mathbf{Y}\mathbf{R}_{\mathbf{1}}^{T}\left(\mathbf{R}_{\mathbf{1}}\mathbf{R}_{\mathbf{1}}^{T}+\epsilon\mathbf{I}_{M}\right)^{-1}$
(by minimizing $\Vert\mathbf{Y}-\mathbf{A}\mathbf{R}_{1}\Vert_{F}^{2}$),
then projected to the valid simplex space as in \cite{zhou2016spatial}.
The initial covariance matrices are set to $\boldsymbol{\Sigma}_{j1}\leftarrow0.1^{2}\mathbf{I}_{B},\,\forall j$.
For the noise matrix $\mathbf{D}$, although there is research focused
on noise estimation \cite{gao2013comparative,roger1996principal},
endmember variability was not considered and validation was performed
only for the simple LMM assumption. Hence, we use an empirical value
$\mathbf{D}=0.001^{2}\mathbf{I}_{B}$, which is usually much less
than the variability of covariance matrices in \eqref{eq:mu_Sigma_nk}.

Step 2: Segmentation. Given the initial conditions, we use the GEM
algorithm to iteratively update $\gamma_{n\mathbf{k}}$ by \eqref{eq:E-step},
$\pi_{\mathbf{k}}$ by \eqref{eq:M-step_wk}, $\boldsymbol{\mu}_{jk}$
by \eqref{eq:M-step_mu_jk1}, $\mathbf{A}$ by \eqref{eq:M-step_A1}
while keeping $\boldsymbol{\Sigma}_{jk}$ fixed. For $\gamma_{n\mathbf{k}}$
and $\pi_{\mathbf{k}}$, a direct update equation is available. For
$\boldsymbol{\mu}_{jk}$, we can use gradient descent. For $\mathbf{A}$,
since we have the non-negativity and sum-to-one constraints, a projected
gradient descent similar to the one used in \cite{zhou2016spatial}
can be applied. To ensure a segmentation effect, a large $\beta_{2}$
is used in this step.

Step 3: Model selection and abundance estimation. Using the segmentation-like
abundance maps from the previous step, we can obtain the interior
pixels $\mathbf{Y}_{j}$ (assumed pure) by thresholding the abundances
(e.g. $\alpha_{nj}>0.99$) and performing image erosion to trim the
boundaries with structure element size $r_{se}$ (can be decreased
gradually if large enough to trim all the pixels). Following Section~\ref{subsec:Model-selection},
we can determine the number of components $K_{j}$ and further calculate
$\boldsymbol{\Theta}_{j}$ by standard EM. Since $\beta_{2}$ is relatively
large in the previous step, it is reduced by $\beta_{2}\leftarrow\zeta\beta_{2}$
where $\zeta=0.05$. Then we restart the optimization to estimate
the abundances with $\boldsymbol{\Theta}$ fixed.

\subsection{Complexity analysis\label{subsec:Complexity-Analysis}}

The abundance estimation algorithm is an iterative process. Since
we used projected gradient descent with adaptive step sizes, the number
of iterations is usually not large as shown in \cite{guan2011manifold,lin2007projected}.
For each iteration, it starts with calculating $\boldsymbol{\mu}_{n\mathbf{k}}$
and $\boldsymbol{\Sigma}_{n\mathbf{k}}$ in \eqref{eq:mu_Sigma_nk},
where storing all $\boldsymbol{\mu}_{n\mathbf{k}}$ ($\boldsymbol{\Sigma}_{n\mathbf{k}}$)
requires $O\left(\left|\mathcal{K}\right|NB\right)$ ($O\left(\text{\ensuremath{\left|\mathcal{K}\right|}}NB^{2}\right)$),
the computation takes $O\left(\text{\ensuremath{\left|\mathcal{K}\right|}}NMB\right)$
($O\left(\left|\mathcal{K}\right|NMB^{2}\right)$). Suppose the Cholesky
factorization and the matrix inversion of a $B$ by $B$ matrix both
take $O\left(B^{3}\right)$ time, and $N\gg B>M$. Evaluating $\log\mathcal{N}\left(\mathbf{y}_{n}\vert\boldsymbol{\mu}_{n\mathbf{k}},\boldsymbol{\Sigma}_{n\mathbf{k}}\right)$
by the Cholesky factorization will take $O\left(B^{3}\right)$, hence
updating all the $\gamma_{n\mathbf{k}}$ takes $O\left(\left|\mathcal{K}\right|NB^{3}\right)$,
which is also the required time for evaluating the objective function
\eqref{eq:obj_fun}. The calculation of $\boldsymbol{\lambda}_{n\mathbf{k}}$,
$\boldsymbol{\Psi}_{n\mathbf{k}}$ (in \eqref{eq:lambda_nk} and \eqref{eq:psi_nk})
will be dominated by the inversion of $\boldsymbol{\Sigma}_{n\mathbf{k}}$
which takes $O\left(B^{3}\right)$, hence the overall calculation
takes $O\left(\left|\mathcal{K}\right|NB^{3}\right)$ with storage
the same as $\boldsymbol{\mu}_{n\mathbf{k}}$ and $\boldsymbol{\Sigma}_{n\mathbf{k}}$.
Then if we move to calculating the derivatives in \eqref{eq:M-step_mu_jk1},
\eqref{eq:M-step_sigma_jk1} and \eqref{eq:M-step_A1}, it is easy
to verify that the computational costs are $O\left(\left|\mathcal{K}\right|NMB\right)$,
$O\left(\left|\mathcal{K}\right|NMB^{2}\right)$, $O\left(\left|\mathcal{K}\right|NMB^{2}\right)$
respectively (Note that $\mathbf{K}$ is a banded matrix so the computation
involving it is linear). Reviewing the above process, we conclude
that the spatial complexity is dominated by $O\left(\left|\mathcal{K}\right|NB^{2}\right)$
and the time complexity is dominated by $O\left(\left|\mathcal{K}\right|NB^{3}\right)$.

\subsection{Estimation of endmembers for each pixel\label{subsec:Estimation-of-endmembers}}

While the previous sections discuss the estimation of the abundances
and endmember distribution parameters, they do not actually estimate
the endmembers $\left\{ \mathbf{m}_{nj}:\,n=1,\dots,N,\,j=1,\dots,M\right\} $
for each pixel. In this Section, we will discuss this additional problem
and note its absence in the previous NCM literature.

Theorem~\ref{thm:linear_comb_GMM2} implies that we can view the
proposed conditional density \eqref{eq:density_y_theta} as modeling
the noise as a Gaussian random variable followed by marginalizing
over $\mathbf{M}_{n}$, which is usually achieved by the evidence
approximation in the machine learning literature due to the intractability
of the integral (Section 3.5 in \cite{bishop2006pattern}). Since
we have $\mathbf{A},\boldsymbol{\Theta}$ obtained from the previous
Sections, we can get the posterior of $\mathbf{M}_{n}$ from this
model:
\begin{align}
p\left(\mathbf{M}_{n}\vert\mathbf{y}_{n},\boldsymbol{\alpha}_{n},\boldsymbol{\Theta},\mathbf{D}\right) & \propto p\left(\mathbf{y}_{n},\mathbf{M}_{n}\vert\boldsymbol{\alpha}_{n},\boldsymbol{\Theta},\mathbf{D}\right)\nonumber \\
 & =p\left(\mathbf{y}_{n}\vert\boldsymbol{\alpha}_{n},\mathbf{M}_{n},\mathbf{D}\right)p\left(\mathbf{M}_{n}\vert\boldsymbol{\Theta}\right).\label{eq:alternate_posterior}
\end{align}
Maximizing $\log p\left(\mathbf{M}_{n}\vert\mathbf{y}_{n},\boldsymbol{\alpha}_{n},\boldsymbol{\Theta},\mathbf{D}\right)$
gives us another minimization problem
\begin{eqnarray}
\mathcal{E}\left(\mathbf{M}_{n}\right) & = & \frac{1}{2}\left(\mathbf{y}_{n}-\mathbf{M}_{n}^{T}\boldsymbol{\alpha}_{n}\right)^{T}\mathbf{D}^{-1}\left(\mathbf{y}_{n}-\mathbf{M}_{n}^{T}\boldsymbol{\alpha}_{n}\right)\nonumber \\
 &  & -\sum_{j=1}^{M}\log\sum_{k=1}^{K_{j}}\pi_{jk}\mathcal{N}\left(\mathbf{m}_{nj}\vert\boldsymbol{\mu}_{jk},\boldsymbol{\Sigma}_{jk}\right)\label{eq:obj_fun_Mn}
\end{eqnarray}
obtained by plugging \eqref{eq:alternate_density_yn} and \eqref{eq:density_Mn}
into \eqref{eq:alternate_posterior}. Note that this objective function
has an intuitive interpretation as the first term minimizes the reconstruction
error while the second term forces the endmembers close to the centers
of each GMM. The weight factor between the two terms is the noise.
From an algebraic perspective, since there are also logarithms of
sums of Gaussian functions in this objective, we can also use the
EM algorithm for ease of optimization. In the E step, the soft membership
is calculated by
\[
\gamma_{njk}=\frac{\pi_{jk}\mathcal{N}\left(\mathbf{m}_{nj}\vert\boldsymbol{\mu}_{jk},\boldsymbol{\Sigma}_{jk}\right)}{\sum_{k}\pi_{jk}\mathcal{N}\left(\mathbf{m}_{nj}\vert\boldsymbol{\mu}_{jk},\boldsymbol{\Sigma}_{jk}\right)},\,k=1,\dots,K_{j}.
\]
In the M step, the derivative w.r.t. $\mathbf{m}_{nj}$ is obtained
as 
\begin{align*}
\frac{\partial\mathcal{E}}{\partial\mathbf{m}_{nj}}= & -\mathbf{D}^{-1}\left(\mathbf{y}_{n}-\mathbf{M}_{n}^{T}\boldsymbol{\alpha}_{n}\right)\alpha_{nj}\\
 & +\sum_{k=1}^{K_{j}}\gamma_{njk}\boldsymbol{\Sigma}_{jk}^{-1}\left(\mathbf{m}_{nj}-\boldsymbol{\mu}_{jk}\right).
\end{align*}
Instead of deploying gradient descent in the M step for estimating
the abundances, combining the derivatives for all \emph{j} actually
leads to a closed form solution
\begin{eqnarray*}
\text{vec}\left(\mathbf{M}_{n}^{T}\right) & = & \left\{ \boldsymbol{\alpha}_{n}\boldsymbol{\alpha}_{n}^{T}\otimes\mathbf{D}^{-1}+\text{diag}\left(\mathbf{C}_{n1},\dots,\mathbf{C}_{nM}\right)\right\} ^{-1}\\
 &  & \left\{ \text{vec}\left(\mathbf{D}^{-1}\mathbf{y}_{n}\boldsymbol{\alpha}_{n}^{T}\right)+\mathbf{d}_{n}\right\} 
\end{eqnarray*}
where $\mathbf{C}_{nj}\in\mathbb{R}^{B\times B}$ and $\mathbf{d}_{n}:=\left(\mathbf{d}_{n1}^{T},\dots,\mathbf{d}_{nM}^{T}\right)^{T}\in\mathbb{R}^{MB\times1}$
are defined as
\[
\mathbf{C}_{nj}:=\sum_{k=1}^{K_{j}}\gamma_{njk}\boldsymbol{\Sigma}_{jk}^{-1},\,\mathbf{d}_{nj}:=\sum_{k=1}^{K_{j}}\gamma_{njk}\boldsymbol{\Sigma}_{jk}^{-1}\boldsymbol{\mu}_{jk}.
\]
In practice, despite the need to estimate a large $M\times B\times N$
tensor, the time cost is actually much less than the estimation of
abundances because of the closed form update equation in the M step.
An interesting fact is that $\gamma_{njk}$ measures the closeness
of estimated endmembers to clusters centers, hence may provide a clue
on which cluster is sampled to generate an endmember.

\section{Results}

In the following experiments, we implemented the algorithm in MATLAB\textsuperscript{\textregistered}
and compared the proposed GMM with NCM, BCM (spectral version with
quadratic programming) \cite{du2014spatial} on synthetic and real
images. As mentioned previously, for GMM, the original image data
were projected to a subspace with 10 dimensions to speed up the computation
for abundance estimation \footnote{The code of GMM is available on GitHub (\url{https://github.com/zhouyuanzxcv/Hyperspectral}).}.
NCM was implemented as a supervised algorithm wherein we input the
ground truth pure pixels (in the image with extreme abundances), modeled
them by Gaussian distributions, and obtained the abundance maps by
maximizing the log-likelihood. We considered two versions of NCM,
one in the same subspace as GMM (referred to as NCM), the other in
the original spectral space (referred to as NCM without PCA). Since
BCM is also a supervised unmixing algorithm, ground truth pure pixels
were again taken as input and the results were the abundance maps.
For GMM and the two versions of NCM, using the algorithm in Section~\ref{subsec:Estimation-of-endmembers}
we can obtain the endmembers for each pixel. All the parameters of
GMM (except the structure element size $r_{se}$) were set to $\beta_{1}=5$,
$\beta_{2}=5$ unless specified throughout the experiments.

For comparison of endmember distributions, we calculated the $L_{2}$
distance $\left(\int\vert f\left(\mathbf{x}\right)-g\left(\mathbf{x}\right)\vert^{2}d\mathbf{x}\right)^{1/2}$
between the fitted distribution and the ground truth one, where the
latter was only available for the synthetic dataset. For comparison
of abundances, we calculated the root mean squared error (RMSE) $\left(\frac{1}{N}\sum_{n}\vert\alpha_{nj}^{GT}-\alpha_{nj}^{est}\vert^{2}\right)^{1/2}$
where $\alpha_{nj}^{GT}$ are the ground truth abundances and $\alpha_{nj}^{est}$
are the estimated values. Since only some pure pixels were identified
as ground truth in the real datasets, we calculated $\text{error}_{j}=\left(\frac{1}{\left|\mathcal{I}\right|}\sum_{n\in\mathcal{I}}\vert\alpha_{nj}^{GT}-\alpha_{nj}^{est}\vert^{2}\right)^{1/2}$
given the pure pixel index set $\mathcal{I}$. For comparison of endmembers,
the same error formula and overall schema were used, i.e. for an index
set $\mathcal{I}_{j}$ of pure pixels for the $j$th endmember (in
the real datasets), $\text{error}_{j}=\frac{1}{\left|\mathcal{I}_{j}\right|}\sum_{n\in\mathcal{I}_{j}}\left(\frac{1}{B}\Vert\mathbf{m}_{nj}^{GT}-\mathbf{m}_{nj}^{est}\Vert^{2}\right)^{1/2}$.

\subsection{Synthetic datasets}

The algorithms were tested for two cases of synthetic images, a supervised
case and an unsupervised case.

\textbf{Supervised}. In this case, a library of ground truth endmembers
were input and the abundances were estimated. The images were of size
$60\times60$ with 103 wavelengths from 430 nm to 860 nm ($\le5$
nm spectral resolution) and created with two endmember classes, meadows
and painted metal sheets, whose spectra were drawn randomly from the
ground truth of the Pavia University dataset (shown in Fig.~\ref{fig:pavia_roi=000026hist},
meadows have 309 samples and painted metal sheets have 941 samples
in the ROI). Since painted metal sheets have multiple modes in the
distribution, it should reflect a true difference between GMM and
the other distributions. The abundances were sampled from a Dirichlet
distribution so each pixel had random values. Also, an additive noise
sampled from $\mathcal{N}\left(\mathbf{n}_{n}|\mathbf{0},\mathbf{D}\right)$
was added to the mixed spectra, where the noise was assumed to be
independent at different wavelengths, i.e. $\mathbf{D}=\text{diag}\left(\sigma_{1}^{2},\dots,\sigma_{B}^{2}\right)$
while $\sigma_{k}$ was again sampled from a uniform distribution
on $\left[0,\,\sigma_{Y}\right]$. 

\begin{figure*}
\begin{centering}
\includegraphics[width=17cm]{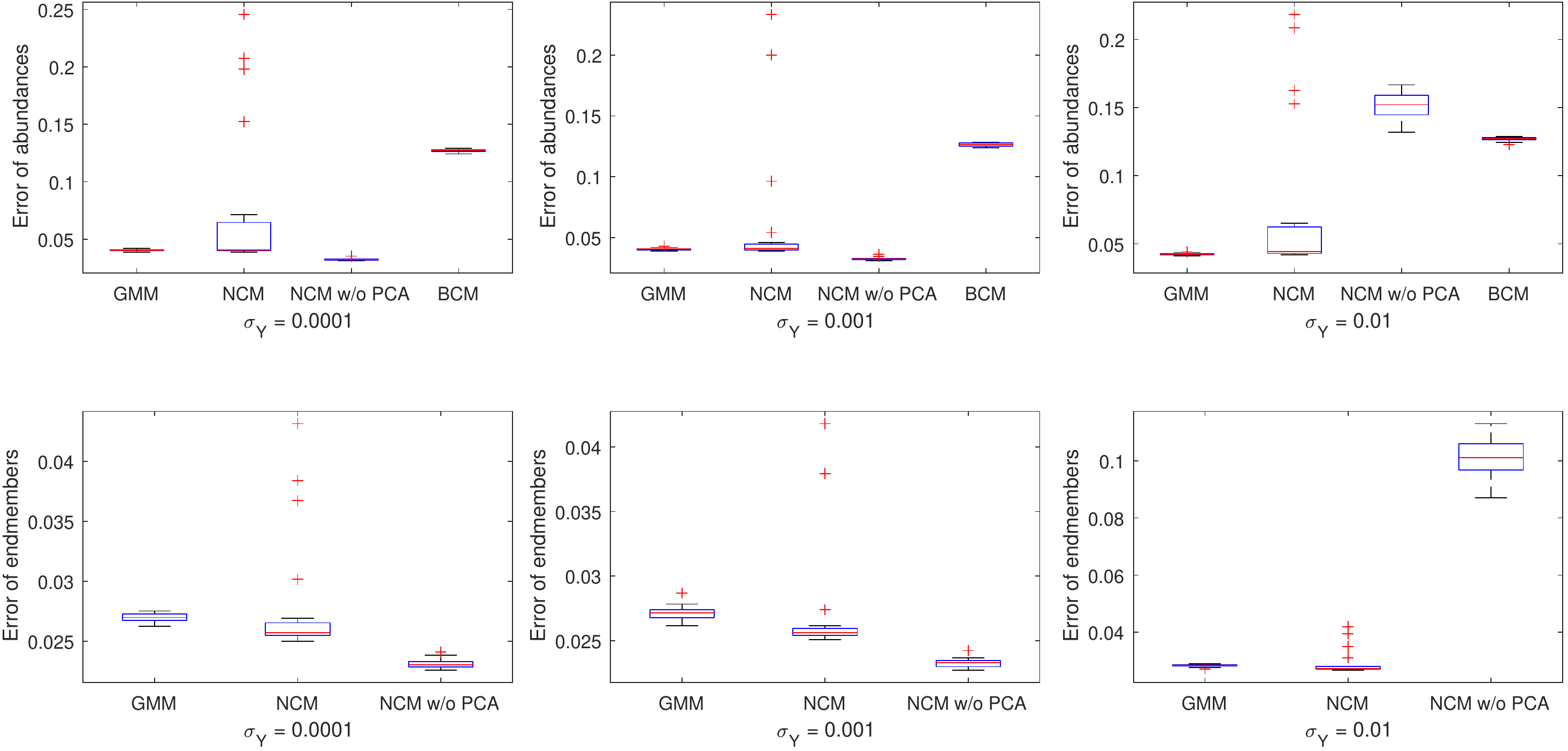}
\par\end{centering}
\caption{Abundance and endmember error statistics from 20 synthetic images
for each noise level in the supervised unmixing scenario.}

\label{fig:abundance_error_supervised}
\end{figure*}

We tested the algorithms for different $\sigma_{Y}$. The effects
of priors were all removed in this case, i.e. $\beta_{1}=0$, $\beta_{2}=0$.
Fig.~\ref{fig:abundance_error_supervised} shows the box plots of
abundance and endmember errors. We can see that GMM has small errors
in general for different noise levels. NCM also has relatively small
errors in most cases, but tends to produce large errors occasionally
(4 out of 20 runs). NCM without PCA has very good results except for
large noise, where it performed worst among all the methods. BCM has
the largest errors overall. For the endmembers, although NCM or NCM
without PCA sometimes has less errors than GMM, the difference is
less than 0.005 hence negligible.

\begin{figure}
\begin{centering}
\includegraphics[width=8.5cm]{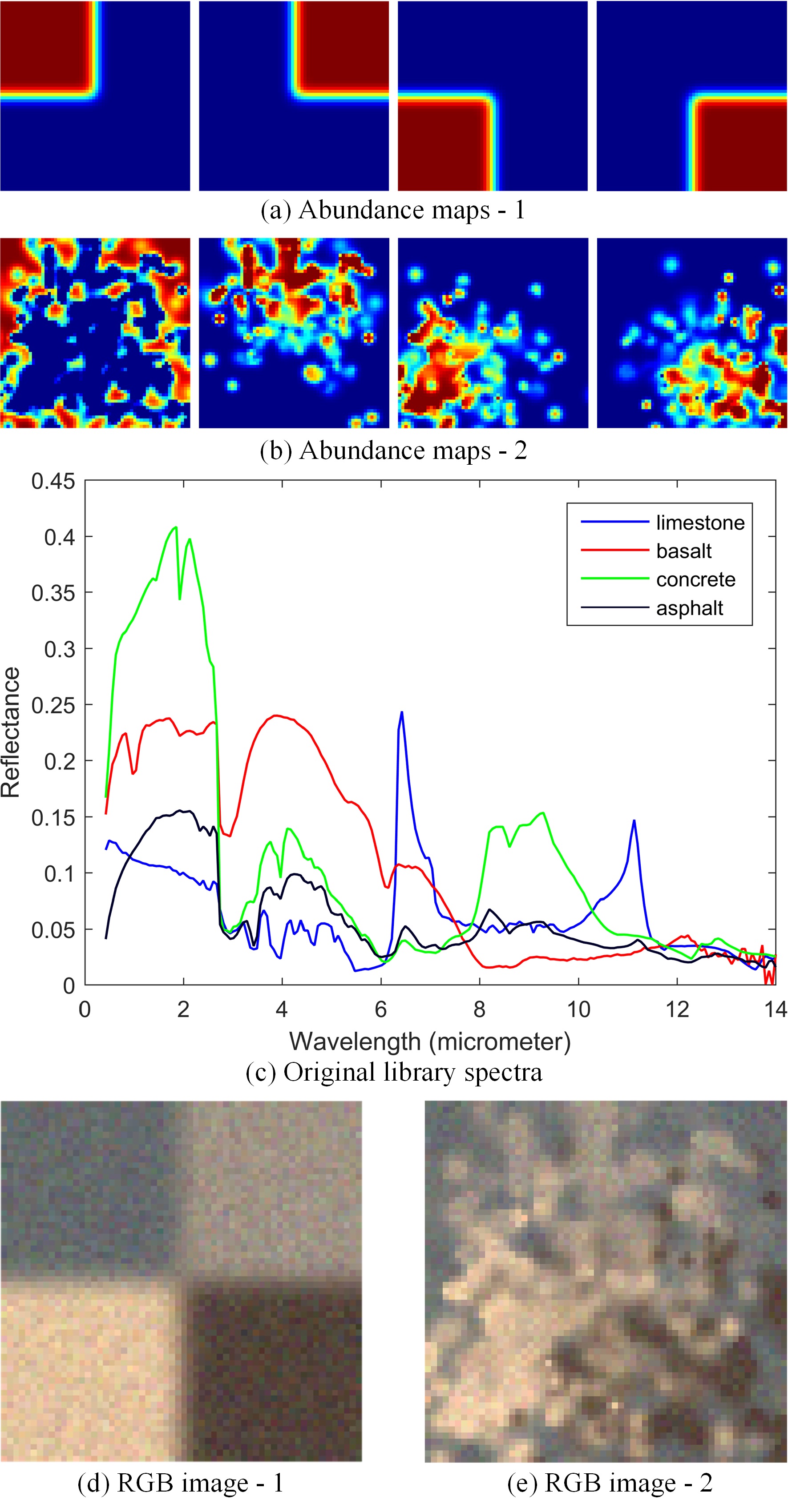}
\par\end{centering}
\caption{Unsupervised synthetic dataset. (a) and (b) are abundance maps for
two images. (c) shows original spectra from the ASTER library. (d)
and (e) show the color images. }

\label{fig:synthetic_data}
\end{figure}

\textbf{Unsupervised}. We created two synthetic images in this case,
the first was used to validate the ability to estimate the distribution
parameters on scenes with regions of pure pixels, the second was used
to validate the segmentation strategy on images with insufficient
pure pixels. They were both of size $60\times60$ pixels and constructed
from 4 endmember classes: limestone, basalt, concrete, asphalt, whose
spectral signatures were highly differentiable. We assumed that the
endmembers were sampled from GMMs following the example in Section~\ref{subsec:An-example}.
The means of the GMMs were from the ASTER spectral library \cite{baldridge2009aster}
(see Fig.~\ref{fig:synthetic_data}(c) for their spectra) with slight
constant changes, which determined a spectral range from 0.4 $\mu$m
to 14 $\mu$m, re-sampled into 200 values. The covariance matrices
were constructed by $a_{jk}^{2}\mathbf{I}_{B}+b_{jk}^{2}\mathbf{u}_{jk}\mathbf{u}_{jk}^{T}$
where $\mathbf{u}_{jk}$ was a unit vector controlling the major variation
direction. For the first image, we assumed the 4 materials occupied
the 4 quadrants of the square image as pure pixels. Then Gaussian
smoothing was applied on each abundance map to make the boundary pixels
of each quadrant be mixed by the neighboring materials. For the second
image, we made the first material as background, the other materials
randomly placed on this background. The procedure of generating the
abundance maps followed \cite{zhou2016spatial}: for each material
(not as background), 150 Gaussian blobs were randomly placed, whose
location and shape width were both sampled from Gaussian distributions.
Finally, noise produced similar to above with $\sigma_{Y}=0.001$
was added to the generated pixels. Fig.~\ref{fig:synthetic_data}
shows the abundance maps, the original spectra of these materials,
and the resulting color images by extracting the bands corresponding
to wavelengths 488 nm, 556 nm, 693 nm.

\begin{figure}
\begin{centering}
\includegraphics[width=8.5cm]{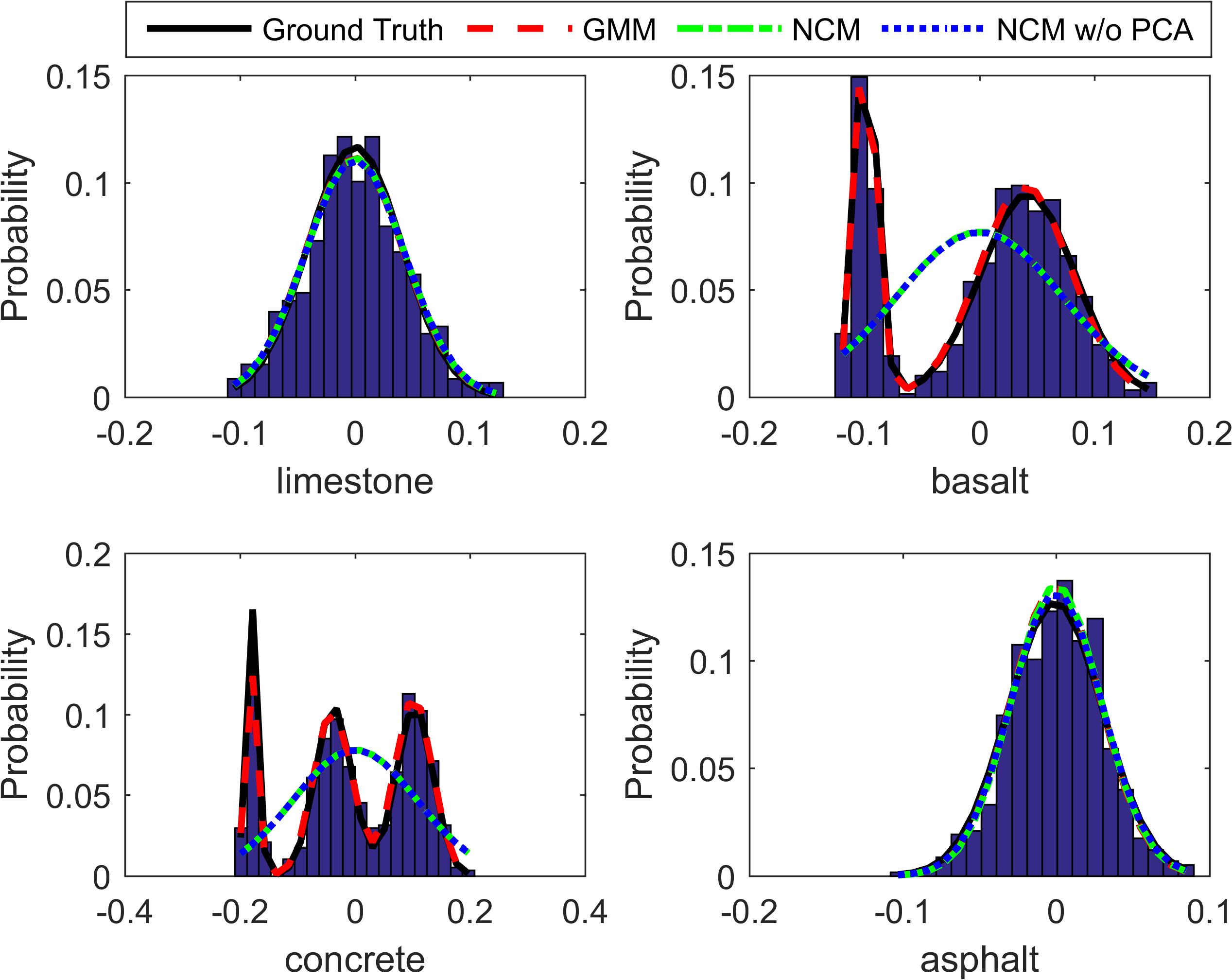}
\par\end{centering}
\caption{Histograms of pure pixels for the 4 materials (when projected to a
1-dimensional space determined by performing PCA on the pure pixels
of each material) and the ground truth and estimated distributions
(also projected to the same direction) for the first image of the
unsupervised synthetic dataset. The probability of each distribution
is calculated by multiplying the value of the density function at
each bin location with the bin size.}

\label{fig:synthetic_histogram}
\end{figure}

The parameters of GMM were $r_{se}=5$ for the two images, $\beta_{1}=0.1$,
$\beta_{2}=0.1$ for the second image. Fig.~\ref{fig:synthetic_histogram}
shows the histograms of ground truth pure pixels and the estimated
distributions for the first image. The ground truth distribution is
barely visible as most of the time it coincides with GMM. For limestone
and asphalt, all the distributions are similar since the pure pixels
are generated by a unimodal Gaussian. However, for basalt and concrete,
GMM provides a more accurate estimation while the two NCMs seem inferior
due to the single Gaussian assumption. The quantitative analysis in
Table~\ref{table:synthetic_dist_diff} implies a similar result by
calculating the $L_{2}$ distance between the estimated distribution
and the ground truth.

\begin{table}
\caption{$L_{2}$ distance between the fitted distributions (GMM, NCM) and
the ground truth distributions for the first image of the unsupervised
synthetic dataset.}

\begin{centering}
\begin{tabular}{|c|c|c|c|c|c|}
\hline 
$\times10^{6}$ &
Limestone &
Basalt &
Concrete &
Asphalt &
Mean\tabularnewline
\hline 
\hline 
GMM &
4.45 &
\textbf{3.46} &
\textbf{3.41} &
4.28 &
\textbf{3.85}\tabularnewline
\hline 
NCM &
\textbf{4.27} &
5.86 &
4.95 &
\textbf{4.02} &
4.77\tabularnewline
\hline 
\end{tabular}
\par\end{centering}
\label{table:synthetic_dist_diff}
\end{table}

Table~\ref{table:synthetic_abundances_error} shows the comparison
of abundance errors from the two images. Since the second image is
much more challenging than the first one, we can expect increased
errors from all the methods. In general, the results of BCM and the
two NCMs show slightly inferior abundances compared to GMM despite
the fact that they have access to pure pixels in the image to train
their models.

\begin{table}
\caption{Abundance errors for the unsupervised synthetic dataset.}

\begin{centering}
\begin{tabular}{|c|c|c|c|c|c|}
\hline 
 &
$\times10^{-4}$ &
GMM &
NCM &
NCM w/o PCA &
BCM\tabularnewline
\hline 
\hline 
\multirow{5}{*}{\begin{turn}{90}
Image 1
\end{turn}} &
Limestone &
\textbf{50} &
107 &
92 &
126\tabularnewline
\cline{2-6} 
 & Basalt &
\textbf{40} &
74 &
67 &
158\tabularnewline
\cline{2-6} 
 & Concrete &
\textbf{41} &
66 &
62 &
186\tabularnewline
\cline{2-6} 
 & Asphalt &
\textbf{69} &
141 &
123 &
292\tabularnewline
\cline{2-6} 
 & Mean &
\textbf{59} &
97 &
86 &
190\tabularnewline
\hline 
\noalign{\vskip0.05cm}
\hline 
\multirow{5}{*}{\begin{turn}{90}
Image 2
\end{turn}} &
Limestone &
\textbf{157} &
1086 &
396 &
231\tabularnewline
\cline{2-6} 
 & Basalt &
\textbf{126} &
445 &
270 &
204\tabularnewline
\cline{2-6} 
 & Concrete &
\textbf{103} &
985 &
229 &
206\tabularnewline
\cline{2-6} 
 & Asphalt &
225 &
\textbf{170} &
706 &
445\tabularnewline
\cline{2-6} 
 & Mean &
\textbf{153} &
671 &
400 &
272\tabularnewline
\hline 
\end{tabular}
\par\end{centering}
\label{table:synthetic_abundances_error}
\end{table}

\subsection{Pavia University}

The Pavia University dataset was recorded by the Reflective Optics
System Imaging Spectrometer (ROSIS) during a flight over Pavia, northern
Italy. The dimension is 340 by 610 with a spatial resolution of 1.3
meters/pixel. It has 103 bands with wavelengths ranging from 430 nm
to 860 nm. As Fig.~\ref{fig:pavia_roi=000026hist} shows, the original
image contains several man-made and natural materials. Considering
that the whole dataset contains many different objects, we only performed
experiments on the exemplar ROI (47 by 106) shown in Fig.~\ref{fig:pavia_roi=000026hist},
in which 5 endmembers, meadows, bare soil, painted metal sheets, shadows
and pavement, are manually identified. 

\begin{figure}
\begin{centering}
\includegraphics[width=8.5cm]{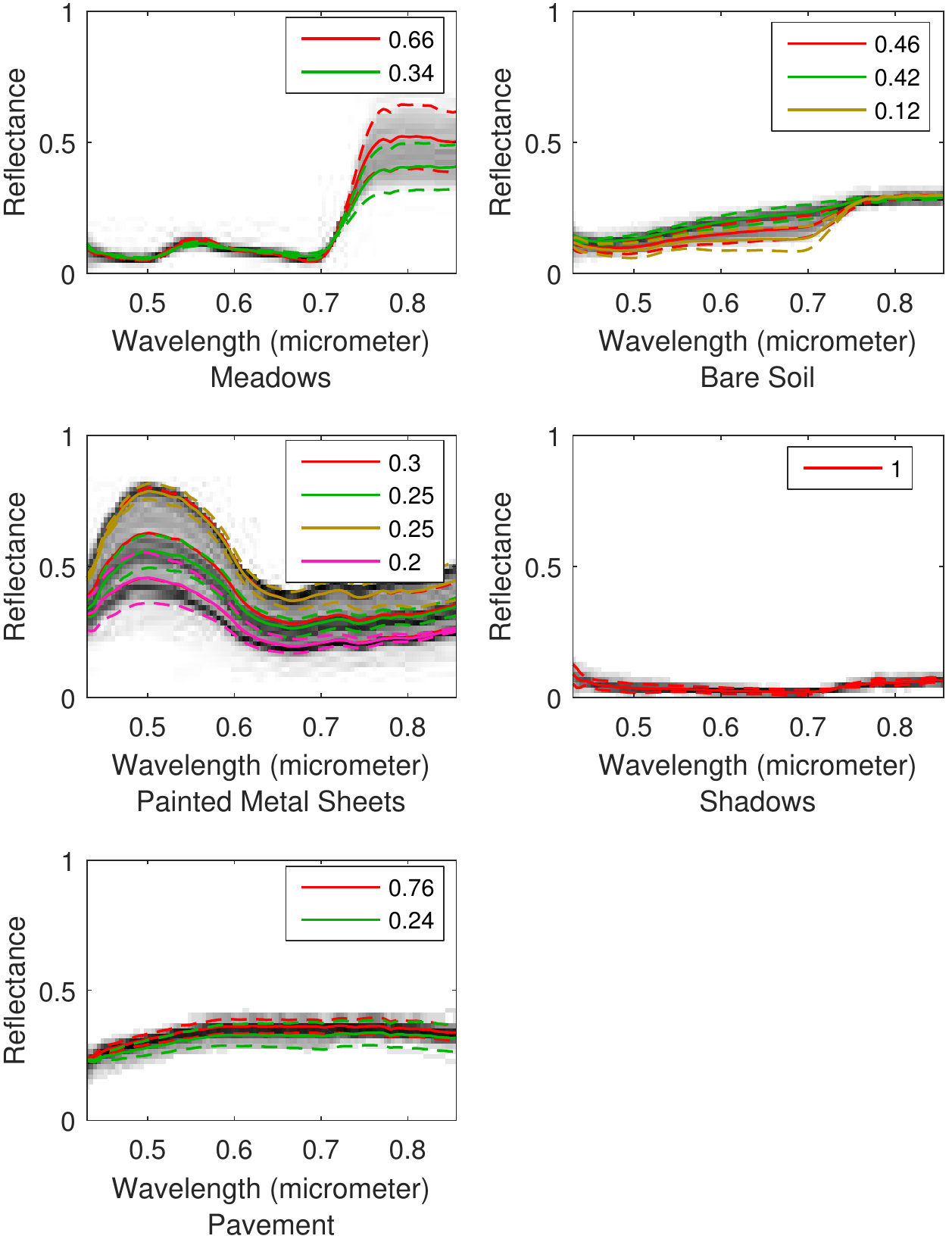}
\par\end{centering}
\caption{Estimated GMM in the wavelength-reflectance space for the Pavia University
dataset. The background gray image represents the histogram created
by placing the pure pixel spectra into the reflectance bins at each
wavelength. The different colors represent different components, where
the solid curve is the center $\boldsymbol{\mu}_{jk}$, the dashed
curves are $\boldsymbol{\mu}_{jk}\pm2\sigma_{jk}\mathbf{v}_{jk}$
($\sigma_{jk}$ is the square root of the large eigenvalue of $\boldsymbol{\Sigma}_{jk}$
while $\mathbf{v}_{jk}$ is the corresponding eigenvector), and the
legend shows the prior probabilities.}

\label{fig:pavia_gmm_wl_reflectance}
\end{figure}

\begin{figure}
\begin{centering}
\includegraphics[width=8.5cm]{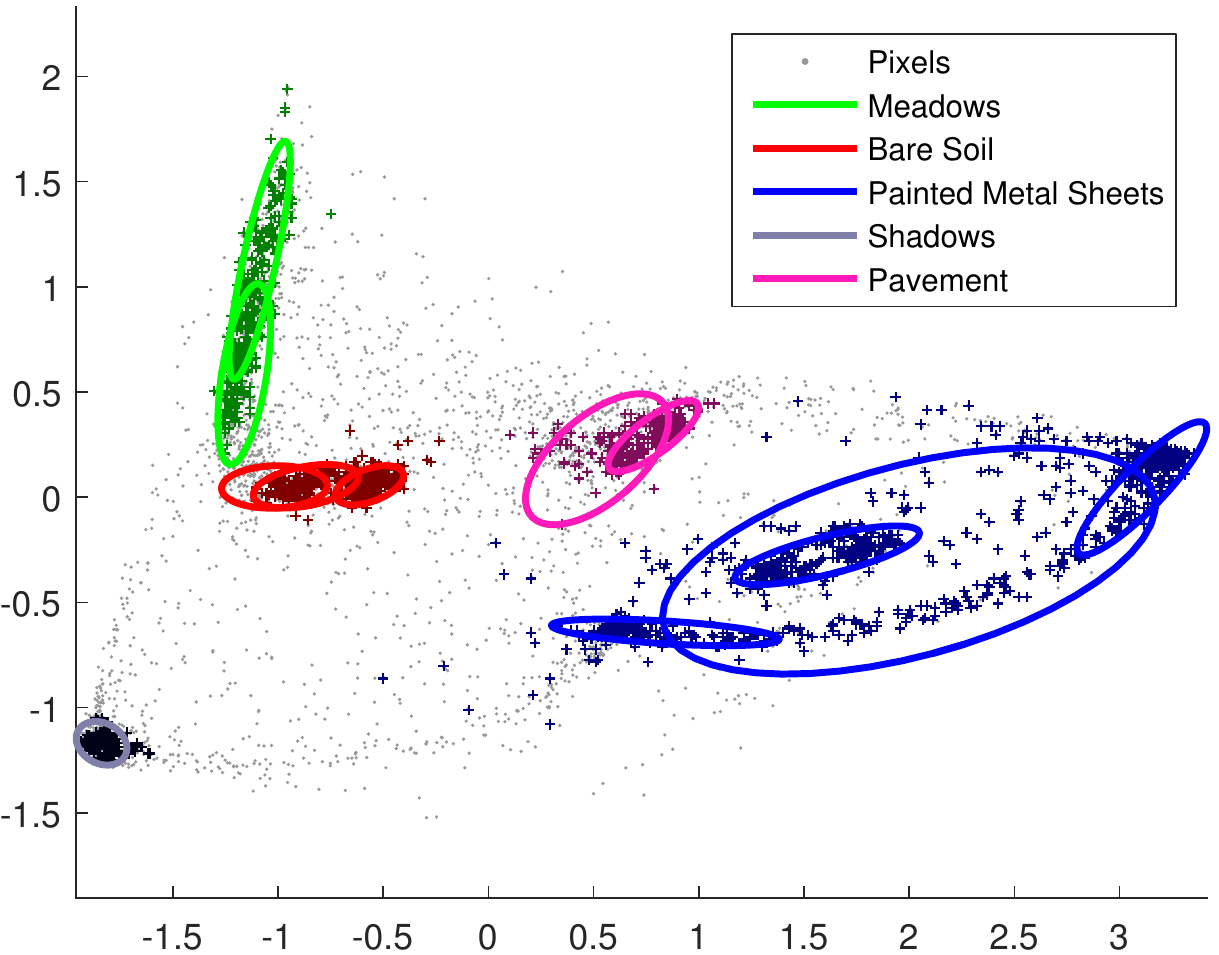}
\par\end{centering}
\caption{Scatter plot of the Pavia University dataset with the estimated GMM.
The gray dots are the projected pixels by PCA. The darkened dots with
a color represent the ground truth pure pixels for a material. The
ellipses with the same color represent the projected Gaussian components
(twice the standard deviation along the major and minor axes, covering
86\% of the total probability mass) for one endmember.}

\label{fig:pavia_gmm_scatter}
\end{figure}

\begin{figure}
\begin{centering}
\includegraphics[width=8.5cm]{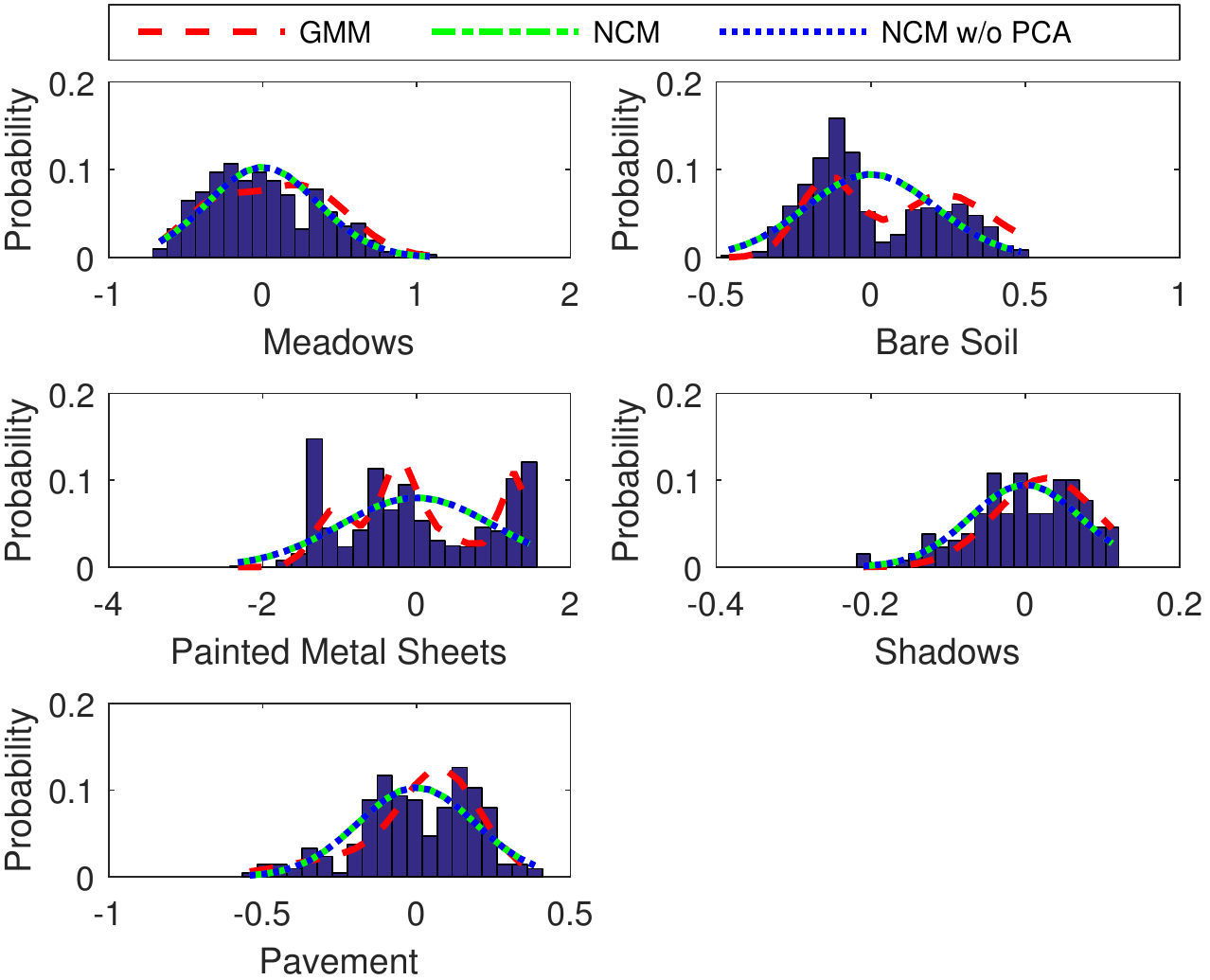}
\par\end{centering}
\caption{Histograms of pure pixels for the Pavia University dataset and the
estimated distributions from GMM and NCM when projected to 1 dimension.}

\label{fig:pavia_histogram}
\end{figure}

The parameter of GMM was $r_{se}=2$. Fig.~\ref{fig:pavia_gmm_wl_reflectance}
shows the GMM in the wavelength-reflectance space, where we can see
the centers and the major variations of the Gaussians. Fig.~\ref{fig:pavia_gmm_scatter}
shows the scatter plot of the results in the projected space. The
scatter plot shows that the identified Gaussian components cover the
ground truth pure pixels very well. For painted metal sheets, which
has a broad range of pure pixels, it estimated 4 components to cover
them. For shadows, only one component was estimated. Fig.~\ref{fig:pavia_histogram}
shows the histograms of pure pixels and the estimated distributions
of GMM and NCMs. We can see that GMM matches the background histogram
better than NCMs.

\begin{figure}
\begin{centering}
\includegraphics[width=8.5cm]{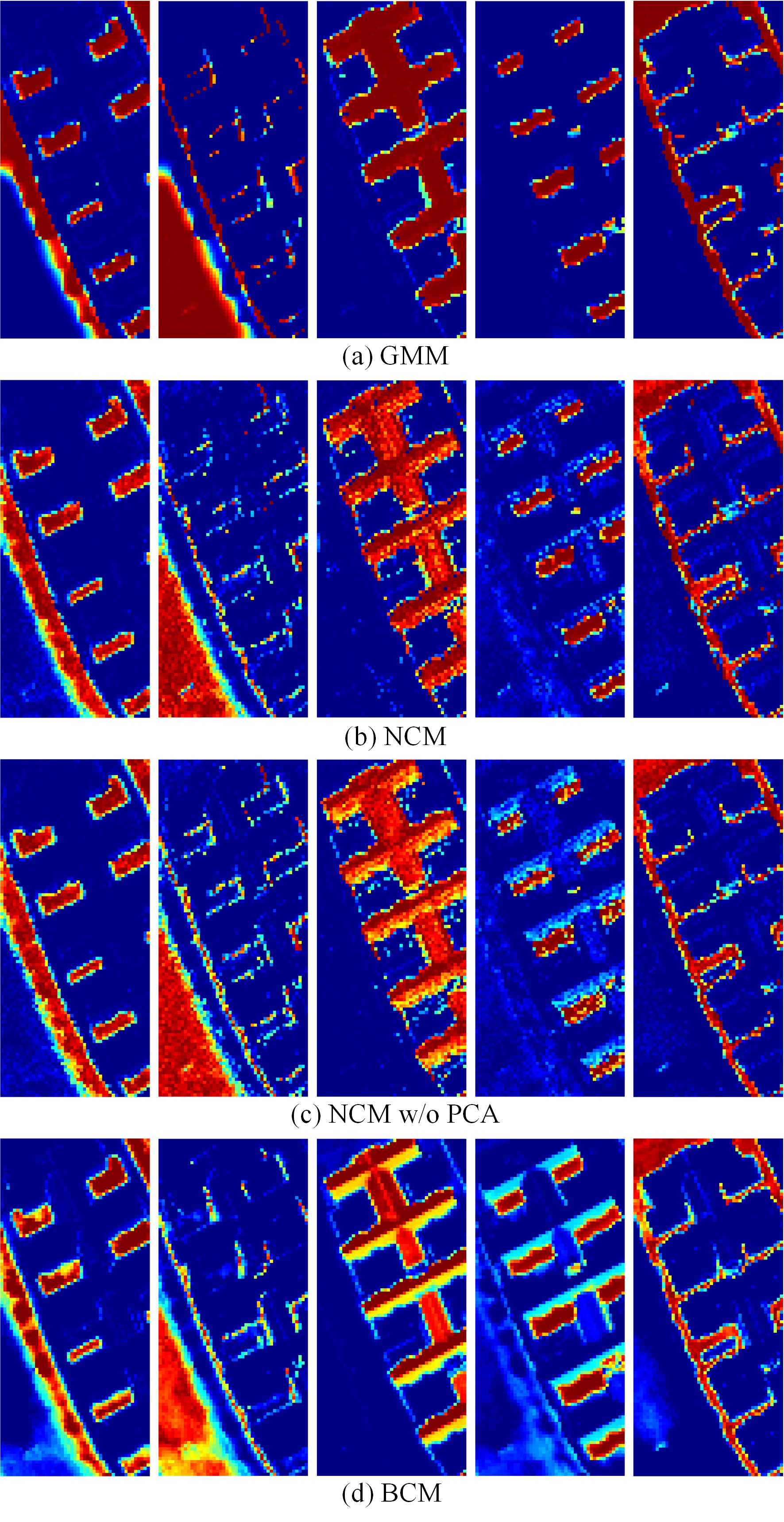}
\par\end{centering}
\caption{Abundance maps for the Pavia University dataset. The corresponding
endmembers from left to right are meadows, bare soil, painted metal
sheets, shadows and pavement.}

\label{fig:pavia_abundances}
\end{figure}

Fig.~\ref{fig:pavia_abundances} shows the abundance map comparison.
Comparing them with the ground truth shown in Fig.~\ref{fig:pavia_roi=000026hist}(a),
we can see that BCM failed to estimate the pure pixels of painted
metal sheets, although ground truth pure pixels were used for training.
For example, the third and fourth abundance maps of BCM show that
the pixels in the lower part of painted metal sheets are mixed with
shadows, while the reduced reflectances are only caused by angle variation.
The result of GMM not only shows sparse abundances for that region,
but also interprets the boundary as a combination of neighboring materials.
Since this dataset has a spatial spacing of 1.3 meters/pixel, we think
this soft transition is more realistic than a simple segmentation.
Although the results of NCMs look good in general, the abundances
in a pure material region are inconsistent. The errors of abundances
and endmembers for these algorithms are shown in Table~\ref{table:pavia_abundances_error},
which implies that GMM performed best overall.

\begin{table}
\centering\caption{Abundance and endmember errors for Pavia University.}

\begin{threeparttable}
\begin{centering}
\begin{tabular}{|c|c|c|c|c|}
\hline 
$\times10^{-4}$ &
GMM &
NCM &
NCM w/o PCA &
BCM\tabularnewline
\hline 
\hline 
Meadow &
\textbf{187} \textbackslash{} \textbf{44}\tnote{a} &
405 \textbackslash{} 113 &
378 \textbackslash{} 114 &
711\tabularnewline
\hline 
Soil &
\textbf{175} \textbackslash{} \textbf{30} &
581 \textbackslash{} 68 &
507 \textbackslash{} 66 &
1049\tabularnewline
\hline 
Metal &
\textbf{476} \textbackslash{} \textbf{49} &
1236 \textbackslash{} 237 &
917 \textbackslash{} 349 &
1285\tabularnewline
\hline 
Shadow &
\textbf{44} \textbackslash{} 44 &
736 \textbackslash{} 48 &
914 \textbackslash{} \textbf{34} &
1287\tabularnewline
\hline 
Pavement &
473 \textbackslash{} \textbf{39} &
1064 \textbackslash{} 114 &
\textbf{333} \textbackslash{} 103 &
612\tabularnewline
\hline 
Mean &
\textbf{271} \textbackslash{} \textbf{41} &
804 \textbackslash{} 116 &
610 \textbackslash{} 133 &
989\tabularnewline
\hline 
\end{tabular}
\par\end{centering}
\label{table:pavia_abundances_error}

\begin{tablenotes} \item [a] the numbers in ".\textbackslash." denote the abundance and endmember errors. \end{tablenotes} \end{threeparttable}
\end{table}

\subsection{Mississippi Gulfport }

\begin{figure}
\begin{centering}
\includegraphics[width=8.5cm]{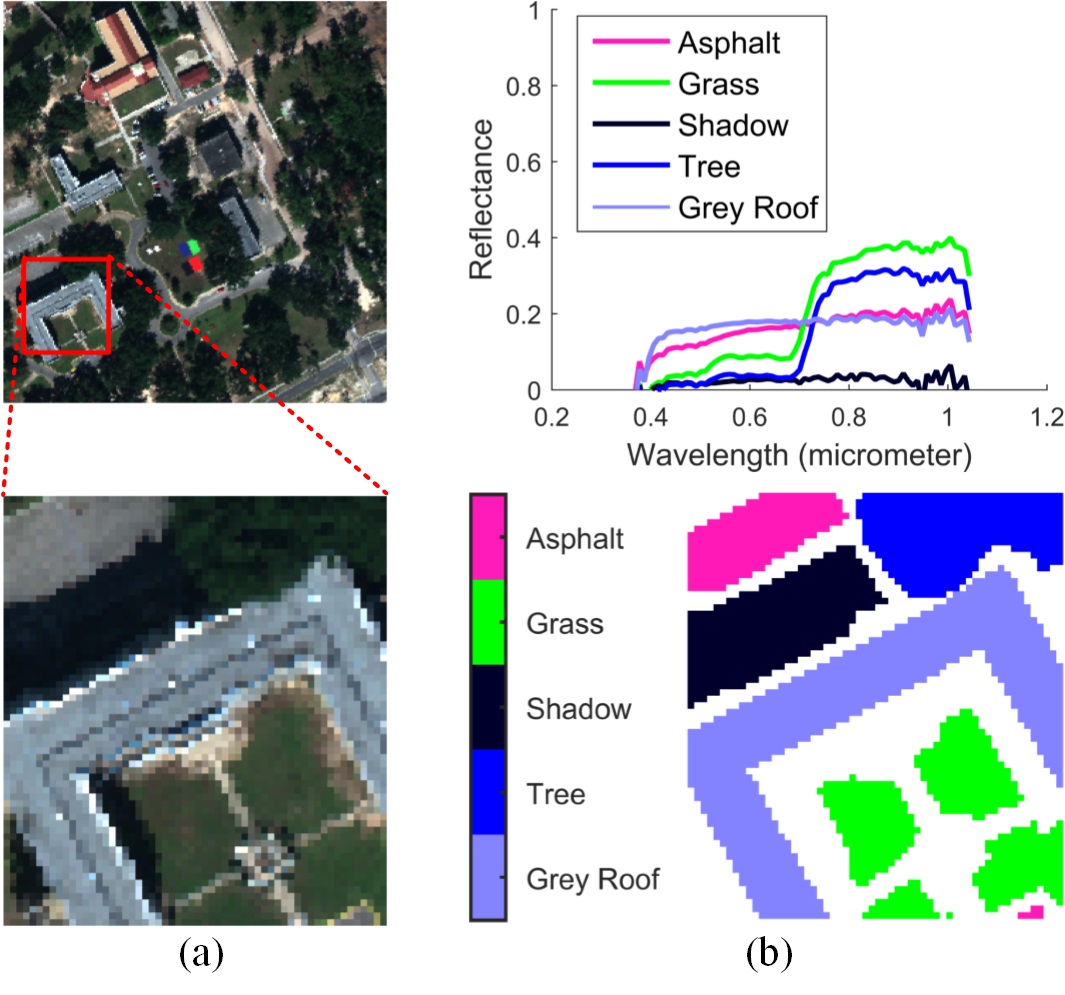}
\par\end{centering}
\caption{(a) Original RGB image of the Mississippi Gulfport dataset with selected
ROI and (b) Ground truth materials in the ROI with their mean spectra.}

\label{fig:gulfport_roi}
\end{figure}

The dataset was collected over the University of Southern Mississippis-Gulfpark
Campus \cite{gader2013muufl}. It is a 271 by 284 image with 72 bands
corresponding to wavelengths 0.368 $\mu m$ to 1.043 $\mu m$. The
spatial resolution is 1 meter/pixel. The scene contains several man-made
and natural materials including sidewalks, roads, various types of
building roofs, concrete, shrubs, trees, and grasses. Since the scene
contains many cloths for target detection, we tried to avoid the cloths
and selected a 58 by 65 ROI that contains 5 materials \cite{Du2017sceneLabel}.
The original RGB image and the selected ROI are shown in Fig.~\ref{fig:gulfport_roi}(a)
while the identified materials and the mean spectra are shown in (b).

\begin{figure}
\begin{centering}
\includegraphics[width=8.5cm]{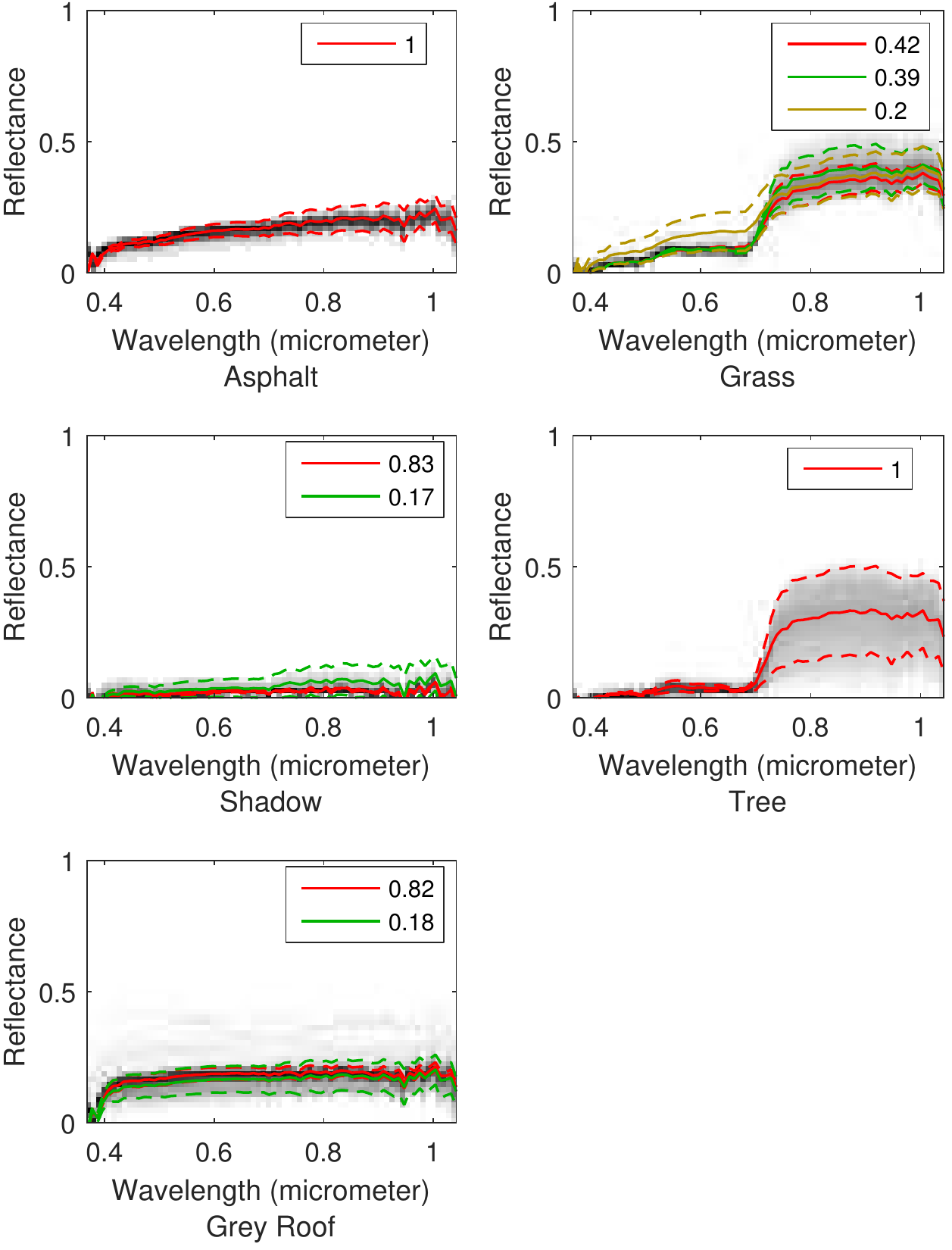}
\par\end{centering}
\caption{Estimated GMM in the wavelength-reflectance space for the Mississippi
Gulfport dataset. The background gray image and the curves have the
same meaning as in Fig.\,\ref{fig:pavia_gmm_wl_reflectance}.}

\label{fig:gulfport_gmm_wl_reflectance}
\end{figure}

\begin{figure}
\begin{centering}
\includegraphics[width=8.5cm]{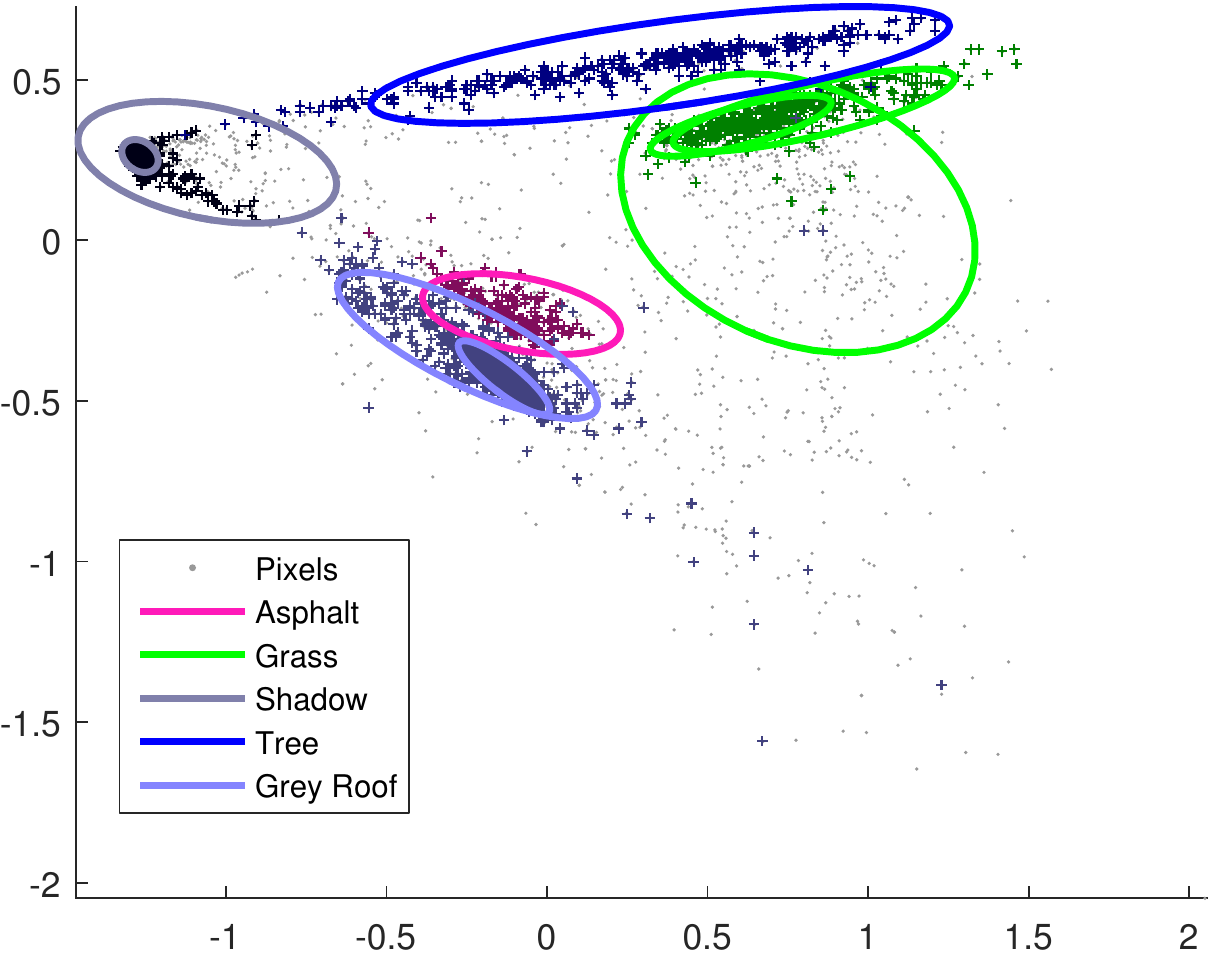}
\par\end{centering}
\caption{Scatter plot of the Mississippi Gulfport dataset with the estimated
GMM. The ellipses and the dots have the same meaning as in Fig.~\ref{fig:pavia_gmm_scatter}.}

\label{fig:gulfport_gmm_scatter}
\end{figure}

\begin{figure}
\begin{centering}
\includegraphics[width=8.5cm]{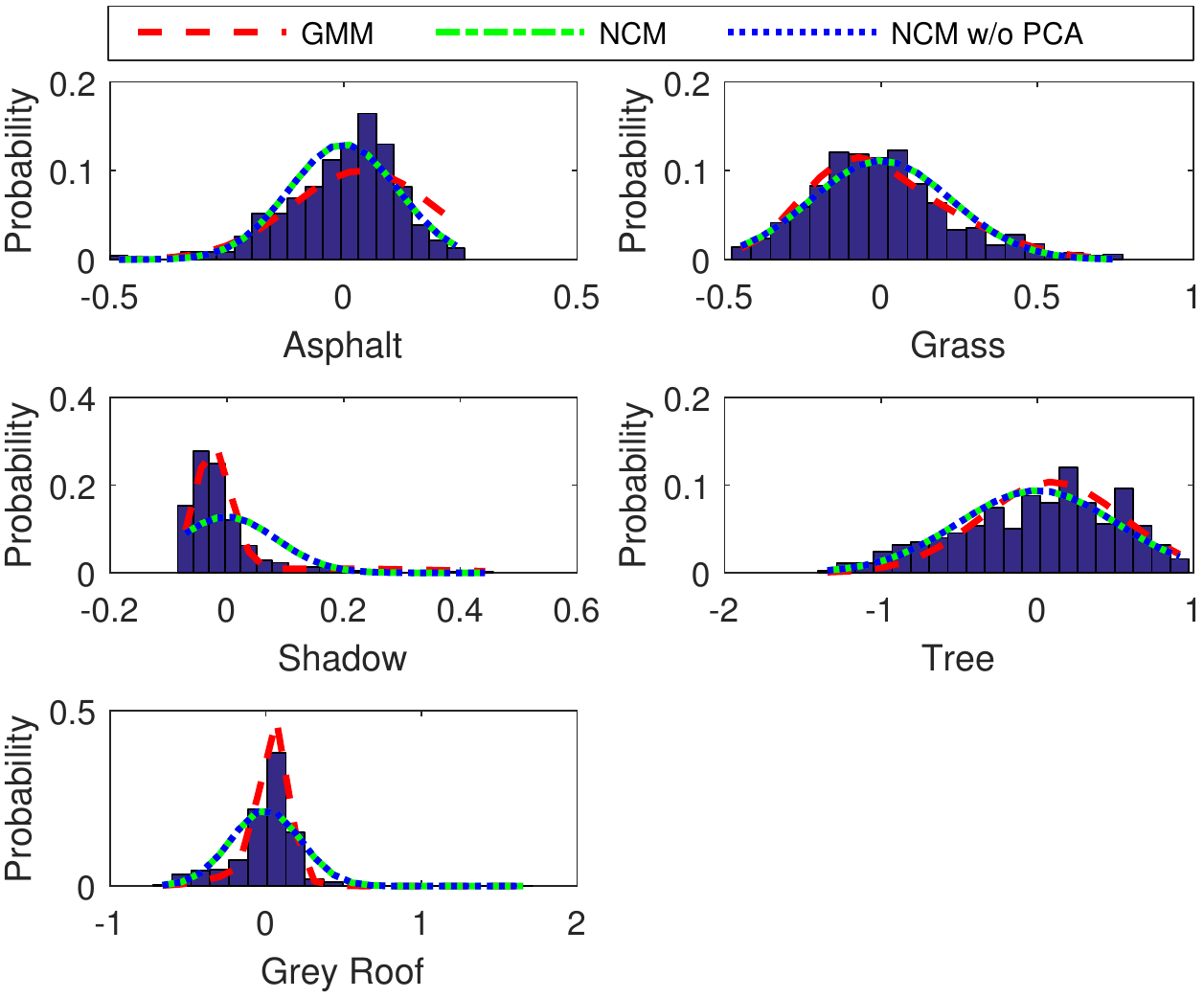}
\par\end{centering}
\caption{Histograms of pure pixels for the Gulfport dataset and the estimated
distributions from GMM and NCM when projected to 1 dimension.}

\label{fig:gulfport_histogram}
\end{figure}

The parameter of GMM was $r_{se}=1$. Fig.~\ref{fig:gulfport_gmm_wl_reflectance}
shows the GMM result in the wavelength-reflectance space and Fig.~\ref{fig:gulfport_gmm_scatter}
shows the scatter plot. We can see that the estimated Gaussian components
successfully cover the identified pure pixels. Fig.~\ref{fig:gulfport_histogram}
shows the estimated distributions. Although there are no multiple
peaks in any of the histograms, NCMs still do not fit the histograms
of shadow and gray roof. In contrast, GMM gives a much better fit
for these 2 endmember distributions.

\begin{figure}
\begin{centering}
\includegraphics[width=8.5cm]{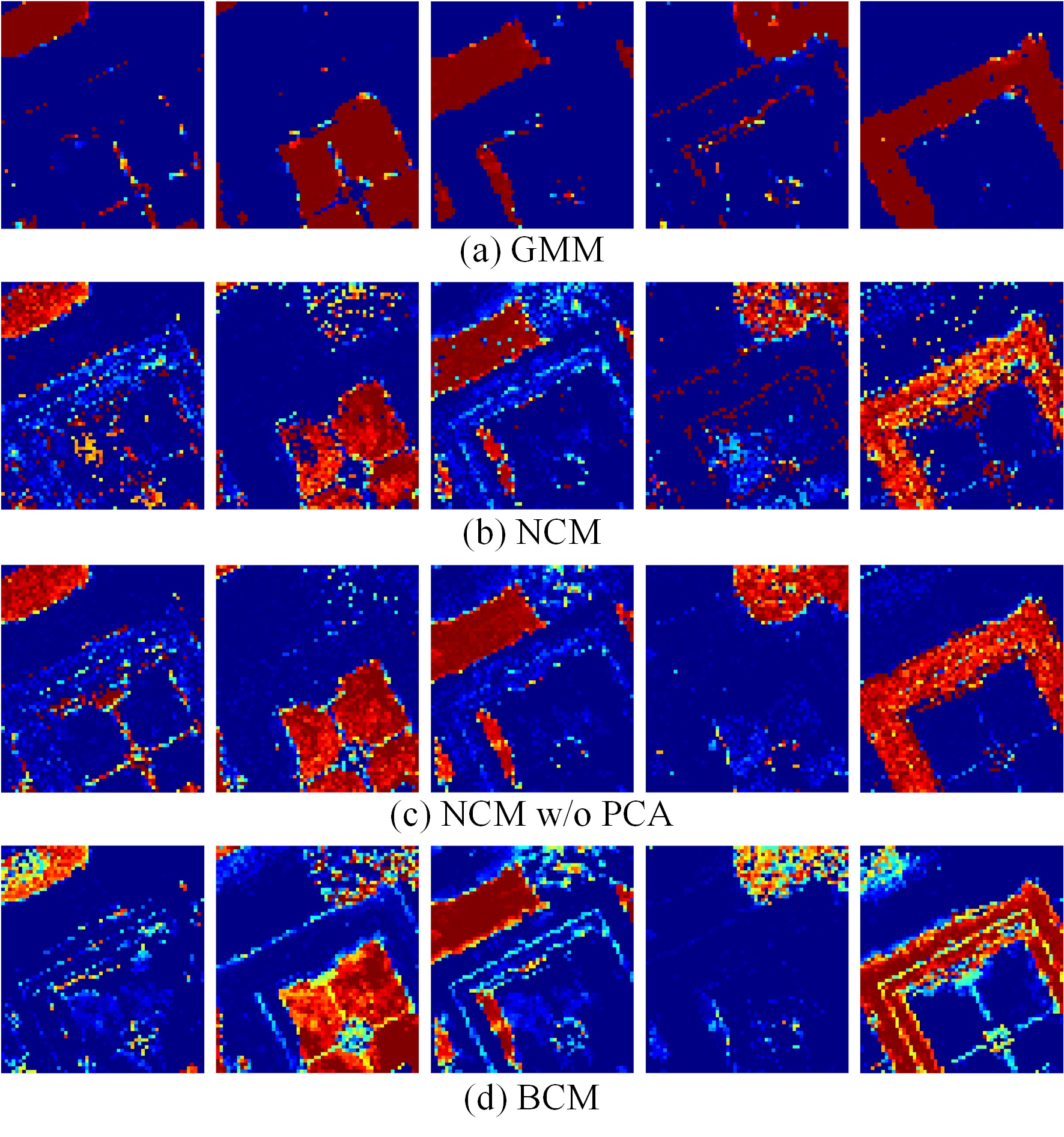}
\par\end{centering}
\caption{Abundance maps for the Gulfport dataset. The corresponding endmembers
from left to right are asphalt, grass, shadow, tree and grey roof.}

\label{fig:gulfport_abundances}
\end{figure}

Fig.~\ref{fig:gulfport_abundances} shows the abundance maps from
different algorithms. We can see that GMM matches the ground truth
in Fig.~\ref{fig:gulfport_roi}(b) best, followed by NCM without
PCA. This is also verified in the quantitative analysis in Table~\ref{table:gulfport_abundances_error}.
Although NCM and BCM take ground truth pure pixels as input, the scattered
dots for trees (fourth abundance map) in both of them and the incomplete
region of grass for NCM (asphalt for BCM) show their insufficiency
in this case.

\begin{table}
\centering

\caption{Abundance and endmember errors for the Gulfport dataset.}

\begin{threeparttable}
\begin{centering}
\begin{tabular}{|c|c|c|c|c|}
\hline 
$\times10^{-4}$ &
GMM &
NCM &
NCM w/o PCA &
BCM\tabularnewline
\hline 
\hline 
Asphalt &
\textbf{205} \textbackslash{}\textbf{ 52}\tnote{a} &
1693 \textbackslash{}\textbf{ }94 &
939 \textbackslash{}\textbf{ }59 &
1420\tabularnewline
\hline 
Grass &
\textbf{169} \textbackslash{} \textbf{58} &
1982 \textbackslash{} 121 &
558 \textbackslash{} 65 &
2145\tabularnewline
\hline 
Shadow &
\textbf{499} \textbackslash{} 49 &
1294 \textbackslash{} 68 &
921 \textbackslash{} \textbf{43} &
1315\tabularnewline
\hline 
Tree &
\textbf{1029} \textbackslash{} \textbf{89} &
2194 \textbackslash{} 234 &
1106 \textbackslash{} 185 &
2279\tabularnewline
\hline 
Roof &
\textbf{908} \textbackslash{} \textbf{76} &
2143 \textbackslash{} 174 &
1234 \textbackslash{} 104 &
1657\tabularnewline
\hline 
Mean &
\textbf{562} \textbackslash{} \textbf{65} &
1861 \textbackslash{} 138 &
952 \textbackslash{} 91 &
1763\tabularnewline
\hline 
\end{tabular}
\par\end{centering}
\label{table:gulfport_abundances_error}

\begin{tablenotes} \item [a] the numbers in ".\textbackslash." denote the abundance and endmember errors. \end{tablenotes} \end{threeparttable}
\end{table}

\section{Discussion and Conclusion}

In this paper, we introduced a GMM approach to represent endmember
variability, by observing that the identified pure pixels in real
applications usually can not be well fitted by a unimodal distribution
as in NCM or BCM. We solved several obstacles in linear unmixing using
this distribution, including (i) deriving the conditional probability
density function of the mixed pixel given each endmember modeled as
GMM from two perspectives; (ii) estimating the abundances and endmember
distributions by maximizing the log-likelihood with a prior enforcing
abundance smoothness and sparsity; (iii) estimating the endmembers
for each pixel given the abundances and distribution parameters. The
results on synthetic and real datasets show superior accuracy compared
to current popular methods like NCM, BCM. Here we have some final
remarks.

\textbf{Complexity}. As analyzed in Section~\ref{subsec:Complexity-Analysis},
each iteration in the estimation of abundances has spatial complexity
$O\left(\left|\mathcal{K}\right|NB^{2}\right)$ and time complexity
$O\left(\left|\mathcal{K}\right|NB^{3}\right)$. For comparison, the
implemented NCM has the same complexity but with $\left|\mathcal{K}\right|=1$.
For the supervised synthetic dataset which contains 60 images, the
total running time of GMM was 9709 seconds, on a desktop with a Intel
Core i7-3820 CPU and 64 GB memory. For comparison, the running time
of NCM, NCM without PCA, and BCM was 941, 50751, 62525 seconds respectively.
In real applications, running GMM on the Pavia University and Mississippi
Gulfport ROIs required 734 seconds and 97 seconds respectively for
abundance estimation (24 seconds and 17 seconds for endmember estimation),
compared to 40 and 34 seconds from NCM, 1389 and 396 seconds from
NCM without PCA, 1170 and 616 seconds from BCM. As analyzed, the main
factors affecting the efficiency of GMM and NCMs are $\left|\mathcal{K}\right|$
and $B$.

\textbf{Limitation}. The complexity analysis leads to one limitation
of the method. That is, the complexity grows exponentially with increasing
numbers of components. This could cause problems for a large amount
of pure pixels. To overcome this shortcoming, there are some empirical
workarounds, such as reducing the number of components by introducing
thresholds, or reducing the number of pure pixels to a fixed number
by random sampling. Another limitation is that the proposed unsupervised
version assumes presence of regions of pure pixels, which mostly happens
in urban scenes. For scenes with a lot of mixed pixels, this assumption
may not hold. Note that unsupervised unmixing is a very challenging
problem. The previous works for this problem all assume several properties
on the abundances and endmembers \cite{drumetz2016blind,thouvenin2016hyperspectral,halimi2016hyperspectral}.
Hence, this limitation exists more or less in all the works on this
problem. Finally, the method was only evaluated on real urban datasets
with only ground truth on pure pixels: it is therefore unclear if
the abundance estimation on mixed pixels is also accurate. This is
due to lack of datasets and ground truth in the hyperspectral community.
We plan to validate it on a more comprehensive dataset given in \cite{wetherley2017mapping}
in the future.

\textbf{Future work}. The proposed GMM formulation has several applications
that we can investigate in the future. First, in target detection,
endmember variability may interfere with the target as well as the
background \cite{jiao2015functions}. By modeling the target or the
background as spectra sampled from GMM distributions, we may devise
more sophisticated and accurate target detection algorithms. Second,
in fusion of hyperspectral and multispectral images, the LMM is usually
used to overcome the underdetermined nature of the problem \cite{yokoya2012coupled,wei2015fast}.
However, the LMM does not hold in real scenarios as shown in this
work. If we use the LMM with endmember variability, which is modeled
by samples from GMM distributions, we may have a fusion algorithm
that better fits the data. Finally, in estimating the noise or intrinsic
dimension of hyperspectral images, simulated data are generated to
quantify the results \cite{gao2013comparative}. When these simulated
data are created, usually the LMM is used without considering the
endmember variability. Using the GMM formulation, we may generate
distinct endmembers for each pixel and create more realistic synthetic
data.

\appendices{}

\bibliographystyle{15C__Users_Yuan_Zhou_Dropbox_YuanHyperspectral_document_GMM_IEEEbib1}

\begin{IEEEbiography}[{\includegraphics[width=1in,height=1.25in]{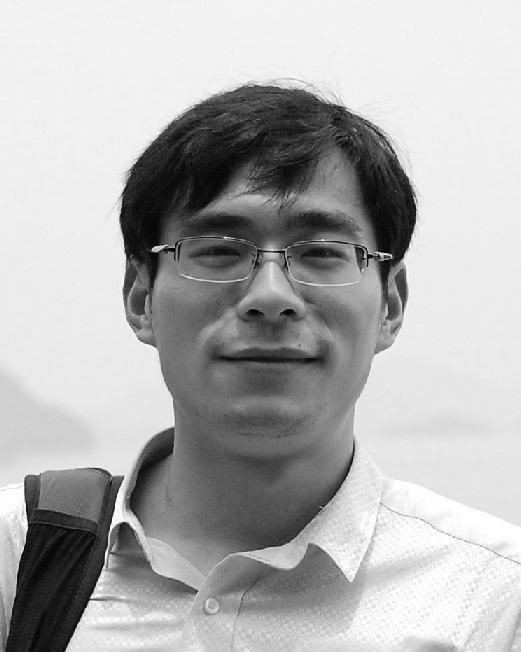}}]{Yuan Zhou}
 received the B.E degree in Software Engineering (2008), the M.E.
degree in Computer Application Technology (2011), both from Huazhong
University of Science and Technology, Wuhan, Hubei, China. Then he
worked in Shanghai UIH as a software engineer for two years. Since
2013, he has been a Ph.D. student in the Department of CISE, University
of Florida, Gainesville, FL, USA. His research interests include image
processing, computer vision and machine learning.
\end{IEEEbiography}

\begin{IEEEbiography}[{\includegraphics[width=1in,height=1.25in]{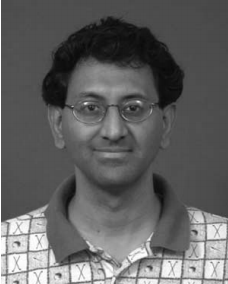}}]{Anand Rangarajan}
 is in the Department of Computer and Information Science and Engineering,
University of Florida, Gainesville, FL, USA. His research interests
are machine learning, computer vision and the scientific study of
consciousness.
\end{IEEEbiography}

\begin{IEEEbiography}[{\includegraphics[width=1in,height=1.25in]{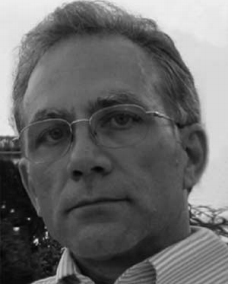}}]{Paul Gader}
 (M\textquoteright 86\textendash SM\textquoteright 09\textendash F\textquoteright 11)
received the Ph.D. degree in mathematics for image-processing-related
research from the University of Florida, Gainesville, FL, USA, in
1986. He was a Senior Research Scientist with Honeywell, a Research
Engineer and a Manager with the Environmental Research Institute of
Michigan, Ann Arbor, MI, USA, and a Faculty Member with the University
of Wisconsin, Oshkosh, WI, USA, the University of Missouri, Columbia,
MO, USA, and the University of Florida, FL, USA, where he is currently
a Professor of Computer and Information Science and Engineering. He
performed his first research in image processing in 1984 working on
algorithms for the detection of bridges in forward-looking infrared
imagery as a Summer Student Fellow at Eglin Air Force Base. He has
since worked on a wide variety of theoretical and applied research
problems including fast computing with linear algebra, mathematical
morphology, fuzzy sets, Bayesian methods, handwriting recognition,
automatic target recognition, biomedical image analysis, landmine
detection, human geography, and hyperspectral and light detection,
and ranging image analysis projects. He has authored/co-authored hundreds
of refereed journal and conference papers.
\end{IEEEbiography}

\end{document}